\documentclass[10pt,twocolumn,letterpaper]{article}

\usepackage[pagenumbers]{cvpr} 

\definecolor{cvprblue}{rgb}{0.21,0.49,0.74}
\usepackage[pagebackref,breaklinks,colorlinks,allcolors=cvprblue]{hyperref}

\def\paperID{*****} 
\def\confName{CVPR}
\def\confYear{2026}
\def\shortTitle{RealDenseFace}

\title{\shortTitle: Real-time Monocular 3D Face Reconstruction from \\Dense UV-space Priors}

\author{
Linzhou Li\quad\quad\quad
Tianjia Shao\quad\quad\quad
Kun Zhou\\
State Key Lab of CAD\&CG, Zhejiang University
}

\begin{document}

\twocolumn[{%
    \renewcommand\twocolumn[1][]{#1}%
    \maketitle
    \begin{center}
      \includegraphics[width=\textwidth]{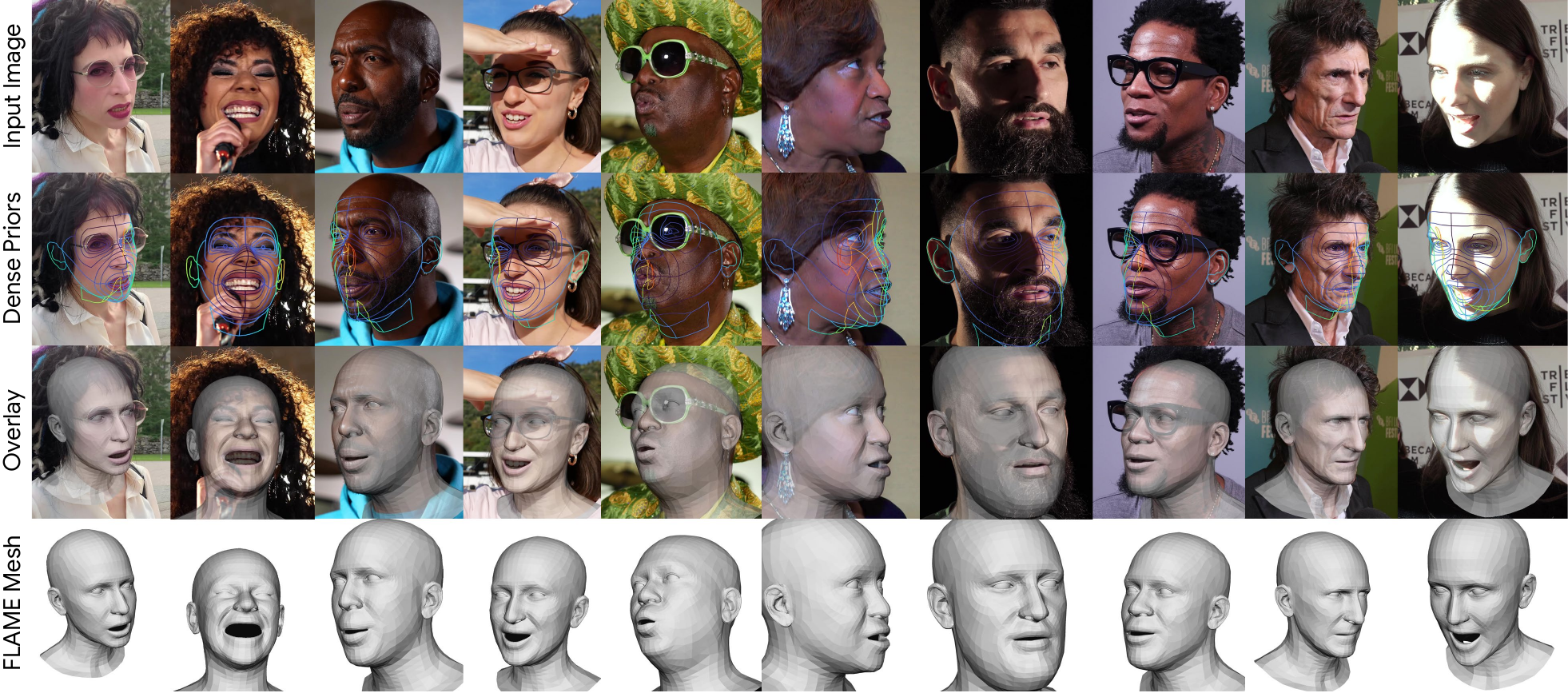}
      \captionof{figure}{ \shortTitle is a real-time optimization-based 3D face reconstruction method with dense UV-space network predictions, achieving real-time online tracking at 84 FPS on an RTX 4090 GPU. We show challenging in-the-wild examples from the VFHQ dataset, covering diverse appearances, poses, expressions, and lightings. From top to bottom: input RGB images, dense prior visualizations, result mesh overlays, and fitted FLAME meshes. The dense prior visualization shows sampled image-space correspondences of selected FLAME vertices, connected by mesh edges and color-coded by the predicted relative depth.}
      \label{fig:in_the_wild}
    \end{center}
}]


\begin{abstract}
Recent monocular 3D face reconstruction methods achieve high fidelity by fitting a 3D Morphable Model (3DMM) to dense priors predicted by networks, but the optimization stage is computationally expensive, often taking tens of seconds per image. We present \shortTitle, a real-time optimization-based 3D face reconstruction method with dense UV-space network predictions. Our key idea is to formulate 3DMM fitting as a nonlinear least-squares problem and solve it with a tailored Gauss-Newton solver that converges in only a few iterations. The reconstruction is conducted in two stages. In the first stage, the network predicts two dense UV-space maps from a single RGB image: a correspondence map for UV-to-image alignment, and a relative-depth map for geometric constraints along the viewing direction. In the second stage, the solver fits per-vertex targets sampled from these maps at the vertex UV coordinates. The solver supports all three reconstruction settings: single-image fitting, offline sequence reconstruction, and online tracking. Our method achieves state-of-the-art accuracy on the NeRSemble SVFR benchmark. The online tracker runs at 80+ FPS, and the offline sequence reconstruction is over 20$\times$ faster than previous optimization-based baselines. Project page: \url{https://gapszju.github.io/RealDenseFace/} 
\end{abstract}

\section{Introduction}
\label{sec:intro}

High-fidelity monocular 3D face reconstruction is fundamental to applications such as telepresence\cite{Face2Face,FaceVR}, facial animation\cite{BouazizOnline13}, and 3D avatar creation\cite{3DGB,RGBAvatar,GaussianAvatars}. Existing methods mainly follow two paradigms: feed-forward regression and optimization-based fitting. Feed-forward methods directly predict parameters of parametric face models such as 3D Morphable Models (3DMMs)\cite{BlanzVetter99,BFM09} with low latency\cite{DECA,MICA,TokenFace,SMIRK,SHeaP}, but they often lag behind optimization-based methods in challenging cases such as large head poses and strong facial expressions\cite{Pixel3DMM}. Optimization-based methods often achieve higher accuracy because they explicitly fit a parametric face model to images at test time~\cite{Face2Face,DDE,FlowFace,Pixel3DMM,WoodDense22,LookMa,MAMMA}, but they are typically slower due to iterative fitting.

Classical optimization-based tracking methods fit parametric face models to sparse landmarks, photometric terms, or optical flow, and some systems achieve real-time monocular tracking\cite{Face2Face,DDE,CaoStabilized18}. However, these signals provide limited dense geometric evidence for high-fidelity reconstruction. Recent methods address this limitation by replacing sparse or photometric cues with dense learned priors. FlowFace\cite{FlowFace} predicts dense UV-to-image correspondences and uses them as 2D alignment priors for FLAME\cite{FLAME} fitting, while Pixel3DMM\cite{Pixel3DMM} predicts screen-space UV coordinates and surface normals to provide additional geometric cues. These methods show that dense priors are highly effective for monocular face reconstruction, but their fitting pipelines commonly rely on a generic optimizer such as Adam. Since such an optimizer requires numerous iterations per frame, high-fidelity fitting with dense priors remains difficult to deploy for real-time reconstruction.


To overcome this limitation, we propose \shortTitle, a real-time face reconstruction framework that performs high-fidelity fitting with dense priors using a tailored Gauss-Newton FLAME optimizer. We formulate FLAME fitting as a compact nonlinear least-squares problem over per-vertex residuals. The FLAME parameters are decomposed into two groups: identity parameters and dynamic parameters, where the latter include expression, local articulation, global head rotation, and translation. For each group, we derive closed-form Jacobians for the residual terms, allowing the solver to update one group while keeping the other fixed. Such groups can then be scheduled differently for different reconstruction settings. For single-image fitting, the optimizer alternates between updating these two groups. For offline sequence reconstruction, we optimize the dynamic parameters of each frame and refine a shared identity over selected keyframes. For online tracking, we track the dynamic parameters of each incoming frame while maintaining an online keyframe buffer to incrementally update the shared identity using only past and current frames.

To make the dense predictions directly usable by our optimizer, we predict dense priors in the FLAME UV domain and sample them at the fixed UV coordinates of FLAME vertices, producing direct vertex-wise fitting targets. We adopt the UV-space formulation instead of screen-space ones~\cite{Pixel3DMM}, because the screen-space fitting is tied to the rendered mesh of each frame and requires sophisticated rasterization-based objective functions during optimization. 
Specifically, our network predicts two dense UV-space priors: a correspondence map and a relative-depth map. The correspondence map provides UV-to-image alignment for FLAME vertices, while the relative-depth map provides geometric constraints along the viewing direction. Compared with correspondence-only priors~\cite{FlowFace}, the relative-depth prior supplies complementary geometric information for resolving depth ambiguity. Rather than predicting absolute depth, we define depth relative to a stable anchor, namely the neck joint. This allows the prior to capture local depth variation on the face while remaining independent of global head translation, which is unreliable under random image cropping. The global translation is then estimated during fitting.

Experiments validate both the accuracy and efficiency of the proposed design. On the official NeRSemble SVFR benchmark\cite{NeRSemble,Pixel3DMM}, our method improves over the state-of-art work Pixel3DMM\cite{Pixel3DMM} on both neutral and posed reconstruction. In terms of runtime, the full single-image pipeline takes \(0.18\,\mathrm{s}\) when camera intrinsics are unknown, where the focal-length search takes \(0.12\,\mathrm{s}\), whereas Pixel3DMM~\cite{Pixel3DMM} takes around \(30\,\mathrm{s}\) per image. To isolate the independent effect of our optimizer, we compare the fitting stage alone: our Gauss-Newton solver converges in \(29.6\,\mathrm{ms}\), while Adam requires \(18\,\mathrm{s}\) under the same fitting setting. For offline sequence reconstruction, our method is over \(24\times\) faster than previous optimization-based baseline VHAP~\cite{VHAP} (i.e., 0.045s versus 1.105s per frame). For online trakcing, our method processes video streams at \(84\) FPS while remaining close to the offline reconstruction accuracy.

Our contributions are summarized as follows:

\begin{itemize}
\item We formulate FLAME fitting from dense priors as a nonlinear least-squares problem and tailor an efficient Gauss-Newton solver for real-time optimization-based reconstruction.
\item We introduce the dense UV-space priors that augment UV-to-image correspondence with relative-depth cues, providing geometric constraints along the viewing direction.
\item We build a unified optimization-based framework that uses the same solver for single-image fitting, offline sequence reconstruction, and online tracking.
\end{itemize}

\begin{figure*}[t]
    \centering
    \includegraphics[width=1.0\linewidth]{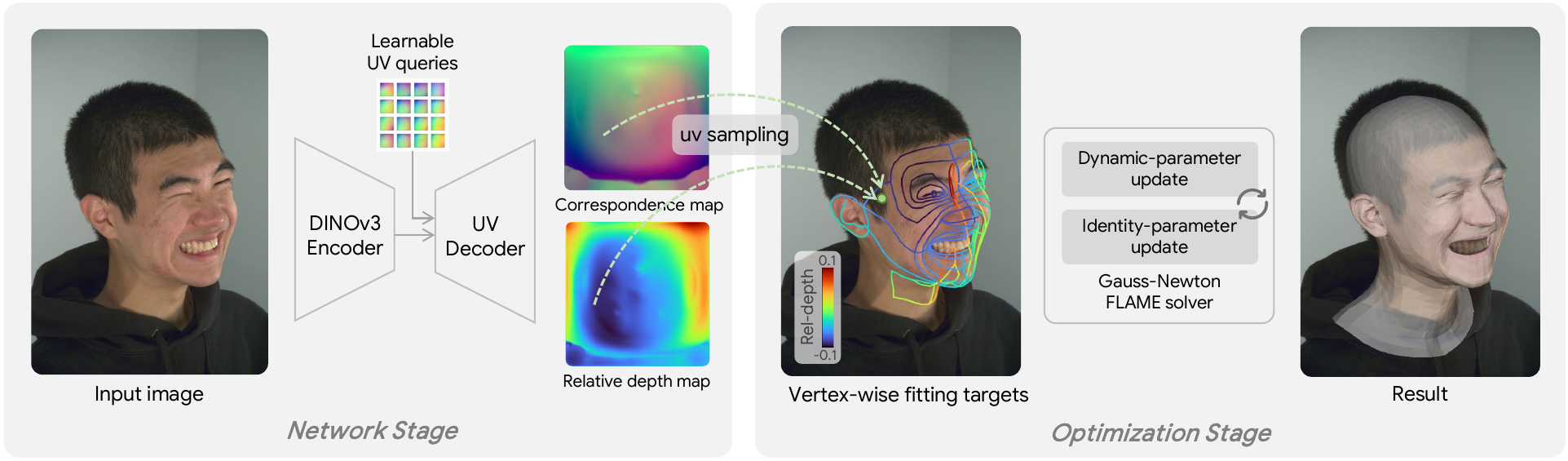}
    \caption{\textbf{Pipeline overview.} Given a portrait image, the network extracts image features with a DINOv3 encoder and decodes them in the FLAME UV domain. It predicts a correspondence map, a relative-depth map, and their log-variance maps (for confidence). These dense UV-space priors are sampled at fixed FLAME vertex UV coordinates to produce vertex-wise 2D alignment targets, relative-depth targets, and confidence weights. The Gauss-Newton solver then alternates between dynamic-parameter updates and identity-parameter updates to fit FLAME identity, expression, pose, and translation parameters by minimizing correspondence and relative-depth residuals. The same solver supports single-image fitting, offline sequence reconstruction, and online tracking.}
    \label{fig:pipeline}
\end{figure*}

\section{Related Work}
\label{sec:related_work}

\subsection{Feed-forward 3D Face Regression}
With the success of deep neural networks in computer vision, many recent methods have focused on feed-forward regression of 3DMM parameters from a single image~\cite{RichardsonSynthetic16,RichardsonDetailed17,Tran3DMM17,NoW,Deep3DFaceRecon19,ThreeDDFAv2,SADRNet,HRN}. These methods avoid test-time optimization and therefore provide low-latency reconstruction. Among them, DECA~\cite{DECA} became a widely used FLAME-based baseline for single-image 3D face reconstruction, and EMOCA~\cite{EMOCA} further improves expressive reconstruction with emotion-aware supervision. TokenFace~\cite{TokenFace} introduces facial component tokens and combines 2D and 3D supervision for more accurate reconstruction. More recent methods improve feed-forward FLAME regression by using more expressive rendering formulations: SMIRK~\cite{SMIRK} introduces analysis-by-neural-synthesis supervision with a neural renderer, while SHeaP~\cite{SHeaP} attaches Gaussians to the predicted head mesh and trains with RGB reconstruction losses. MICA~\cite{MICA} focuses on metrical neutral identity reconstruction rather than full expression reconstruction. Despite their efficiency, feed-forward methods often remain less accurate than optimization-based fitting under challenging poses and strong expressions~\cite{FlowFace,Pixel3DMM}.

\subsection{Classical Optimization-based Methods}
Parametric face models~\cite{BlanzVetter99,BFM09,FaceWarehouse,FLAME} provide a strong prior for monocular 3D face reconstruction by constraining the ill-posed image-to-geometry problem to a low-dimensional space of identity, expression, and pose parameters. Early real-time facial performance capture systems used RGBD input to obtain robust geometric evidence for tracking~\cite{WeisePerformance11,LiHaoCorrectives13,BouazizOnline13,ThiesReenactment15}. Later methods extended face tracking to monocular RGB input~\cite{CaoTOG13,DDE,CaoTOG15,Face2Face,CaoStabilized18}. Among them, DDE~\cite{DDE} uses sparse landmarks for real-time expression tracking, Face2Face~\cite{Face2Face} relies on photometric consistency, and Cao et al.~\cite{CaoStabilized18} combine optical flow with a learned dynamic rigidity prior to improve tracking stability. More recent fitting frameworks, including Metrical Tracker~\cite{MetricalTracker} and VHAP~\cite{VHAP}, improve reconstruction quality with a more expressive parametric face model, i.e., FLAME~\cite{FLAME}, and appearance modeling, but are designed as offline optimization pipelines. In contrast to these optimization-based fitting pipelines, our method uses learned dense priors instead of sparse landmarks, photometric terms, or optical flow for parametric model fitting.

\subsection{Dense Priors for Optimization-based Methods}
Beyond direct feed-forward regression, another line of work improves reconstruction quality by using networks to predict intermediate priors for parametric model fitting, while retaining the flexibility of optimization-based reconstruction. Wood et al.~\cite{WoodDense22} and Chandran et al.~\cite{ContinuousLandmark} predict dense landmarks for parametric face model fitting. More recent methods predict richer dense priors: FlowFace~\cite{FlowFace} predicts UV-to-image alignment for 3D face tracking, Pixel3DMM~\cite{Pixel3DMM} predicts screen-space UV and normal priors for single-image fitting, and Pix2NPHM~\cite{Pix2NPHM} explores test-time refinement for a more expressive neural parametric head model~\cite{NPHM}. These methods improve reconstruction quality, but their fitting stages rely on first-order optimization and are designed mainly for offline reconstruction. In contrast, we formulate dense-prior fitting as a nonlinear least-squares problem and solve it with a tailored Gauss-Newton optimizer, making optimization-based reconstruction much faster.


\section{Method}
\label{sec:method}

Given a portrait image, our method reconstructs a FLAME~\cite{FLAME} face model through a two-stage pipeline (Fig.~\ref{fig:pipeline}): dense-prior prediction followed by Gauss-Newton FLAME fitting. The network predicts a UV-to-image correspondence map and a relative-depth map in UV space, and the fitting stage uses the obtained vertex-wise targets to optimize FLAME parameters.
In the following, we describe the dense UV-space priors in Section~\ref{sec:prior}, present the tailored Gauss-Newton FLAME solver in Section~\ref{sec:solver}, and explain how the solver is used for single-image fitting, offline sequence reconstruction, and online tracking.

\subsection{Dense UV-space Priors}
\label{sec:prior}

We predict dense priors in UV space so that the network output can be used directly as targets for FLAME fitting. Given a portrait image \(I\) as input, the prediction network \(f_{\theta}\) outputs a correspondence map \(\mathbf{C}\), a relative-depth map \(\mathbf{D}\), and their log-variance maps \(\mathbf{L}^{\mathrm{c}}\) and \(\mathbf{L}^{\mathrm{d}}\):
\begin{equation}
(\mathbf{C}, \mathbf{L}^{\mathrm{c}}, \mathbf{D}, \mathbf{L}^{\mathrm{d}}) = f_{\theta}(I).
\end{equation}
Here the correspondence map \(\mathbf{C}: \mathcal{U} \rightarrow \mathbb{R}^2\) maps a UV location to its image-space coordinate, and the relative-depth map \(\mathbf{D}: \mathcal{U} \rightarrow \mathbb{R}\) predicts a scalar relative-depth value for each UV location. At test time, we sample these maps at the pre-defined UV coordinate of each FLAME vertex. The sampled correspondence and relative depth serve as vertex-wise targets for FLAME fitting, while the sampled log-variances provide their per-vertex confidence weights. Thus, the predicted priors can directly be used as least-squares residual targets.

\noindent \textbf{Prior definition.} The correspondence prior provides a dense 2D alignment target for each projected FLAME vertex. However, 2D correspondence alone may not solve the depth ambiguity. We therefore introduce a relative-depth prior to provide complementary geometric constraints for each vertex.

The relative-depth prior is defined with respect to a stable FLAME anchor. Specifically, we use the FLAME neck joint as the reference point. Although named the neck joint, this joint is located near the center of the head in the FLAME model rather than at the literal human neck, and its position is nearly unaffected by facial expression. It therefore provides a stable anchor for defining relative facial depth. For FLAME vertex \(i\), the relative-depth target is
\begin{equation}
d_i = z_i^{\mathrm{cam}} - z_{\mathrm{neck}}^{\mathrm{cam}},
\end{equation}
where \(z_i^{\mathrm{cam}}\) and \(z_{\mathrm{neck}}^{\mathrm{cam}}\) denote the camera-space depths of vertex \(i\) and the neck joint. We use relative rather than absolute depth because the random cropping and resizing used during training make the absolute scale inconsistent across images, leaving absolute depth an unstable prediction target. By subtracting the anchor depth, the prior removes the global translation component and focuses on the relative 3D geometry of the face. The global translation is then recovered in the fitting stage by optimization.

\noindent \textbf{Network architecture.} We instantiate \(f_{\theta}\) with a DINOv3~\cite{DINOv3} image encoder and two UV-space decoder branches. The encoder extracts multi-scale features from the input image. A learnable UV feature map serve as queries tied to the FLAME UV layout. Each decoder branch fuses these UV queries with the image features and produces a UV-aligned feature map. We use separate branches for correspondence and relative depth. Each branch is followed by a DPT head that predicts the target map and its log-variance map.

\noindent \textbf{Training.} Following FlowFace~\cite{FlowFace}, we supervise the predictions at two levels. The dense UV loss supervises the whole UV map, while the vertex loss supervises the per-vertex values. For both levels, we use Gaussian negative log-likelihood loss~\cite{GNLL}. Let \(q \in \{\mathrm{c}, \mathrm{d}\}\) denote either correspondence or relative depth. For UV pixel \(p\), let \(y_p^q\), \(\hat{y}_p^q\), and \(\hat{\ell}_p^q\) denote the ground-truth target, predicted target, and predicted log-variance. The dense loss is
\begin{equation}
\mathcal{L}_{\mathrm{dense}}^{q} =
\frac{1}{|\mathcal{U}|}\sum_{p \in \mathcal{U}} w_p
\left(
\frac{1}{2}\hat{\ell}_p^q +
\frac{1}{2}\left\| y_p^q - \hat{y}_p^q \right\|_2^2
\exp(-\hat{\ell}_p^q)
\right),
\label{eq:dense_loss}
\end{equation}
where \(w_p\) denotes different weights for different head areas in UV space. Please refer the supplementary document for details. 

We then sample the predicted maps at the FLAME vertex UV coordinates and apply the same loss to these per-vertex predictions. For sampled vertex target \(y_i^q\), prediction \(\hat{y}_i^q\), and log-variance \(\hat{\ell}_i^q\), we define
\begin{equation}
\mathcal{L}_{\mathrm{vert}}^{q} =
\frac{1}{N_v}\sum_i w_i
\left(
\frac{1}{2}\hat{\ell}_i^q +
\frac{1}{2}\left\| y_i^q - \hat{y}_i^q \right\|_2^2
\exp(-\hat{\ell}_i^q)
\right),
\label{eq:vertex_loss}
\end{equation}
where \(w_i\) denotes the vertex weights. Please see the supplementary material for details. The full training loss is
\begin{equation}
\mathcal{L} =
\mathcal{L}_{\mathrm{vert}}^{\mathrm{c}} +
\lambda_{\mathrm{dense}} \mathcal{L}_{\mathrm{dense}}^{\mathrm{c}} +
\lambda_{\mathrm{d}} (
\mathcal{L}_{\mathrm{vert}}^{\mathrm{d}} +
\lambda_{\mathrm{dense}} \mathcal{L}_{\mathrm{dense}}^{\mathrm{d}}
).
\end{equation}
In practice, we use a shared dense weight \(\lambda_{\mathrm{dense}} = 0.01\), and set the depth weight \(\lambda_{\mathrm{d}} = 8.0\) to balance the different numerical scales of the correspondence and relative-depth targets.

\subsection{Tailored Gauss-Newton FLAME Solver}
\label{sec:solver}

\noindent We solve FLAME fitting as a nonlinear least-squares problem over the dense priors predicted in Section~\ref{sec:prior}. Previous dense-prior fitting methods~\cite{Pix2NPHM,FlowFace} rely on a generic optimizer such as Adam, which require hundreds of iterations for convergence. In contrast, our objective is explicitly a sum of squared residuals, making it well suited to Gauss-Newton optimization, which converges in far fewer iterations than Adam. We derive close-form Jacobians and implement the solver with custom CUDA kernels.

\noindent \textbf{Parameter grouping.} We group FLAME parameters into identity parameters and dynamic parameters. The identity coefficients are denoted by \(\beta \in \mathbb{R}^{300}\). For frame \(t\), the dynamic parameters are denoted as
\begin{equation}
\mathbf{x}_t = [\boldsymbol{\psi}_t,\ \boldsymbol{\theta}_t,\ \boldsymbol{\phi}_t,\ \mathbf{t}_t] \in \mathbb{R}^{118},
\end{equation}
where \(\boldsymbol{\psi}_t \in \mathbb{R}^{100}\) denotes expression, \(\boldsymbol{\theta}_t \in \mathbb{R}^{3}\) is the global rotation, \(\boldsymbol{\phi}_t \in \mathbb{R}^{12}\) is the local articulation parameters, and \(\mathbf{t}_t \in \mathbb{R}^{3}\) is the global translation. Given \((\beta, \mathbf{x}_t)\), FLAME produces posed vertices and joints
\begin{equation}
(\mathbf{V}_t, \mathbf{J}_t) = \mathrm{FLAME}(\beta, \mathbf{x}_t).
\end{equation}
During optimization, we alternate between updating the dynamic parameters with identity fixed and updating the identity parameters with the dynamic parameters fixed.

\noindent \textbf{Vertex-wise targets from dense priors.} For each FLAME vertex \(i\), we sample the predicted correspondence, relative-depth, and log-variance maps at its pre-defined UV coordinate \(\nu_i\), yielding\(
\hat{\mathbf{u}}_{t,i},\ \hat{d}_{t,i},\ \hat{\ell}^{\mathrm{c}}_{t,i},\ \hat{\ell}^{\mathrm{d}}_{t,i}.
\) These sampled quantities are fixed targets during fitting. Given the current FLAME parameters and camera, we compute the projected 2D vertex position and the relative depth
\begin{equation}
\mathbf{u}_{t,i} = \Pi(\mathbf{V}_{t,i}), \qquad
d_{t,i} = z^{\mathrm{cam}}_{t,i} - z^{\mathrm{cam}}_{t,\mathrm{neck}},
\end{equation}
where \(\Pi\) denotes the camera projection and \(z^{\mathrm{cam}}_{t,\mathrm{neck}}\) is the camera-space depth of the FLAME neck joint. Fitting then minimizes the difference between the targets \((\hat{\mathbf{u}}_{t,i}, \hat{d}_{t,i})\) and the current FLAME estimates \((\mathbf{u}_{t,i}, d_{t,i})\).

We define the weighted correspondence and relative-depth residuals as
\begin{equation}
\begin{aligned}
\mathbf{r}^{\mathrm{c}}_{t,i}
&=
\sqrt{\lambda_{\mathrm{c}}}\,
\exp\!\left(-\frac{1}{2}\hat{\ell}^{\mathrm{c}}_{t,i}\right)
\bigl(\mathbf{u}_{t,i} - \hat{\mathbf{u}}_{t,i}\bigr), \\
r^{\mathrm{d}}_{t,i}
&=
\sqrt{\lambda_{\mathrm{d}}}\,
\exp\!\left(-\frac{1}{2}\hat{\ell}^{\mathrm{d}}_{t,i}\right)
\bigl(d_{t,i} - \hat{d}_{t,i}\bigr).
\end{aligned}
\end{equation}
Here \(\lambda_{\mathrm{c}}\) and \(\lambda_{\mathrm{d}}\) balance the correspondence and relative-depth terms, while the predicted log-variances provide vertex-wise confidence weights. Vertices with higher predicted uncertainty contribute less to the fitting objective.

The dynamic parameters are optimized by minimizing
\begin{equation}
E_{\mathbf{x}_t}
=
\sum_i
\left(
\|\mathbf{r}^{\mathrm{c}}_{t,i}\|_2^2
+
\|r^{\mathrm{d}}_{t,i}\|_2^2
\right)
+
\lambda_{\mathrm{expr}}\|\boldsymbol{\psi}_t\|_2^2
+
\lambda_{\mathrm{pose}}\|\boldsymbol{\phi}_t\|_2^2.
\end{equation}
The identity parameters are optimized from the same data terms accumulated over multiple frames:
\begin{equation}
E_{\beta}
=
\sum_{t,i}
\left(
\|\mathbf{r}^{\mathrm{c}}_{t,i}\|_2^2
+
\|r^{\mathrm{d}}_{t,i}\|_2^2
\right)
+
\lambda_{\mathrm{id}}\|\beta - \beta_{\mathrm{mica}}\|_2^2.
\end{equation}
The weights \(\lambda_{\mathrm{expr}}\), \(\lambda_{\mathrm{pose}}\), and \(\lambda_{\mathrm{id}}\) control the strengths of the expression, pose, and identity regularization terms, respectively. Here (\(\beta_{\mathrm{mica}}\)) is the identity coefficient predicted by MICA~\cite{MICA} from the input image, and the identity term keeps the optimized (\(\beta\)) close to the MICA estimate.

\begin{figure*}[!t]
    \centering
    \includegraphics[width=0.85\linewidth]{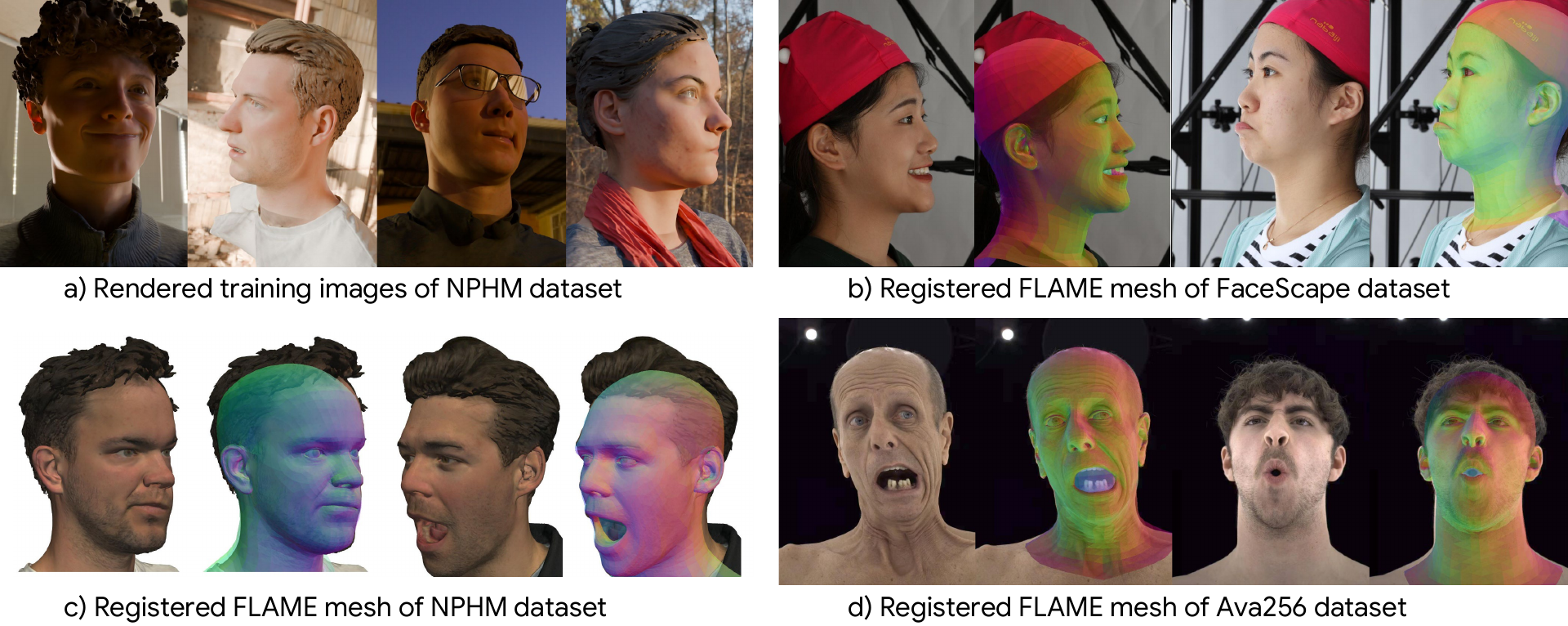}
    \caption{\textbf{Dataset preprocessing.} Examples of the preprocessing used to construct supervision for dense-prior training. For NPHM, we render textured scans under randomized viewpoints and HDRI lighting to obtain RGB training images. For FaceScape~\cite{FaceScape}, NPHM, and AVA-256~\cite{Ava256}, we obtain FLAME-topology registrations aligned to the available scan or tracked geometry.}

    \label{fig:dataset}
\end{figure*}

\noindent \textbf{Gauss-Newton optimization.} For each group update, we linearize the current residual vector \(\mathbf{r}\) with respect to the active variables and solve the damped Gauss-Newton system
\begin{equation}
(\mathbf{J}^{\top}\mathbf{J} + \lambda_{\mathrm{GN}} \mathbf{I}) \Delta = -\mathbf{J}^{\top}\mathbf{r},
\end{equation}
where \(\mathbf{J}\) is the Jacobian matrix and \(\lambda_{\mathrm{GN}}\) is the damping coefficient. The active variables are then updated by
\begin{equation}
\mathbf{x}_t \leftarrow \mathbf{x}_t + \Delta_{\mathbf{x}}, \qquad
\beta \leftarrow \beta + \Delta_{\beta}.
\end{equation}
For dynamic-parameter updates, \(\mathbf{J}\) contains derivatives with respect to \(\mathbf{x}_t\). For identity-parameter updates, \(\mathbf{J}\) contains derivatives with respect to \(\beta\). We solve the resulting normal equations with Cholesky factorization. More details are provided in the supplementary material.

\subsection{Single-Image and Sequence Fitting}
\label{sec:fitting}

\noindent We use the optimizer in Section~\ref{sec:solver} under three settings: single-image fitting, offline sequence reconstruction, and online tracking. These settings share the same vertex-wise targets, fitting energy, and Gauss-Newton optimization, and differ only in how the update of dynamic-parameter and identity-parameter is scheduled.

\noindent \textbf{Single-image fitting.} For a single image, we initialize the identity (\(\beta\)) with the MICA prediction (\(\beta_{\mathrm{mica}}\)), and first optimize only the global pose parameters \(\{\boldsymbol{\theta}, \mathbf{t}\}\) while keeping expression and identity fixed. This pose-only stage gives a stable initialization and reduces the risk of poor local minima in the subsequent full fitting. Starting from this initialization, we run group descent between the dynamic parameters \(\mathbf{x}\) and the identity parameters \(\beta\): each iteration updates \(\mathbf{x}\) by minimizing \(E_{\mathbf{x}}\) with \(\beta\) fixed, and then updates \(\beta\) by minimizing \(E_{\beta}\) on the same frame with \(\mathbf{x}\) fixed.

\noindent \textbf{Offline sequence reconstruction.} For a monocular sequence, we first apply the single-image procedure to the first frame to initialize \(\beta\) and \(\mathbf{x}_0\). We then fix \(\beta\) and track the full sequence in temporal order. For each frame, \(\mathbf{x}_t\) is initialized as the optimized \(\mathbf{x}_{t-1}\) and refined by minimizing \(E_{\mathbf{x}_t}\). Given the fitting results, we select keyframes by farthest point sampling to cover diverse facial expressions and poses, and use these keyframes to refine the shared identity. Specifically, we run group descent on the selected keyframes. With the refined identity, we run another full tracking pass over the sequence. We repeat this keyframe identity refinement and full-sequence tracking for three rounds.

\noindent \textbf{Online tracking.} In the online setting, the input is a video stream, so frames are processed in temporal order without access to future frames. The first frame is fitted as above. It is then inserted into a fixed-size keyframe buffer and used to initialize the shared identity. For each subsequent frame, we initialize \(\mathbf{x}_t\) as the previous dynamic parameters \(\mathbf{x}_{t-1}\), keep the current identity fixed, and update only the dynamic parameters by minimizing \(E_{\mathbf{x}_t}\).

To refine identity online, we maintain a keyframe buffer that stores past frames with diverse head rotations. Every several frames, we compute the head rotation of the current frame as \(R^{\mathrm{head}}_t = R^{\mathrm{root}}_t R^{\mathrm{neck}}_t\). We measure its minimum geodesic distance on \(\mathrm{SO}(3)\) to the head rotations already stored in the buffer. If the buffer is not full, the candidate is inserted only when this distance exceeds a threshold. If the buffer is full, the candidate is inserted by replacing an existing keyframe only when this improves the head pose coverage of the keyframe buffer.

Whenever the keyframe buffer is updated, we add a small identity-refinement budget rather than immediately running a full keyframe optimization. This budget is consumed over subsequent frames: at most one group-descent iteration is applied per frame. This avoids sudden computation spikes, while still allowing the shared identity \(\beta\) to improve as more diverse keyframes become available.

\begin{figure*}[!t]
    \centering
    \includegraphics[width=0.85\linewidth]{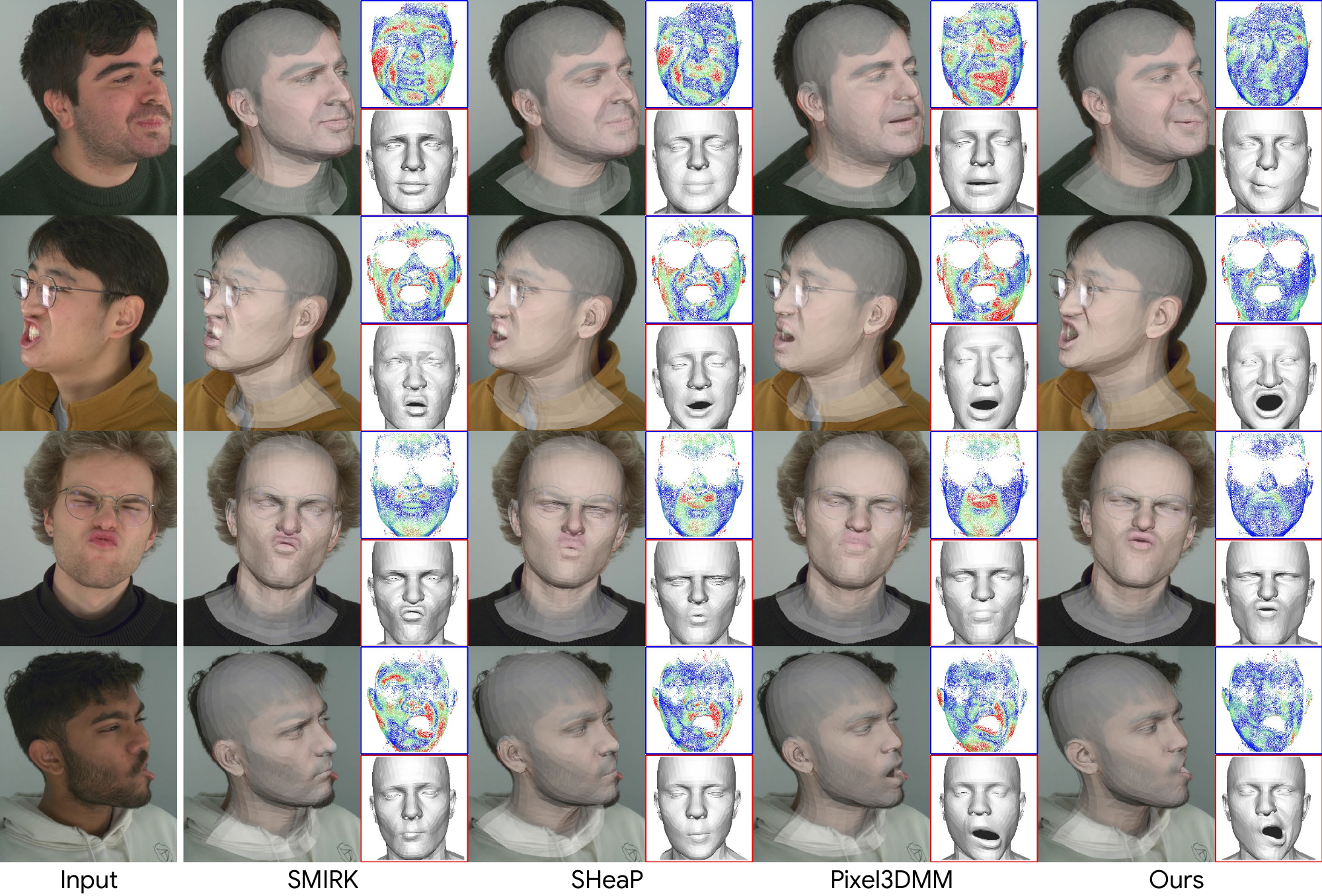}
    \caption{\textbf{Qualitative comparison on NeRSemble SVFR benchmark (posed).} Our method remains accurate under large head pose and strong facial expressions, while SHeaP~\cite{SHeaP} and Pixel3DMM~\cite{Pixel3DMM} tend to under-fit the expression, and SMIRK~\cite{SMIRK} often fails to reconstruct the correct expression under large poses. }
    \label{fig:comp_nersemble}
\end{figure*}

\begin{figure*}[!t]
    \centering
    \includegraphics[width=1.0\linewidth]{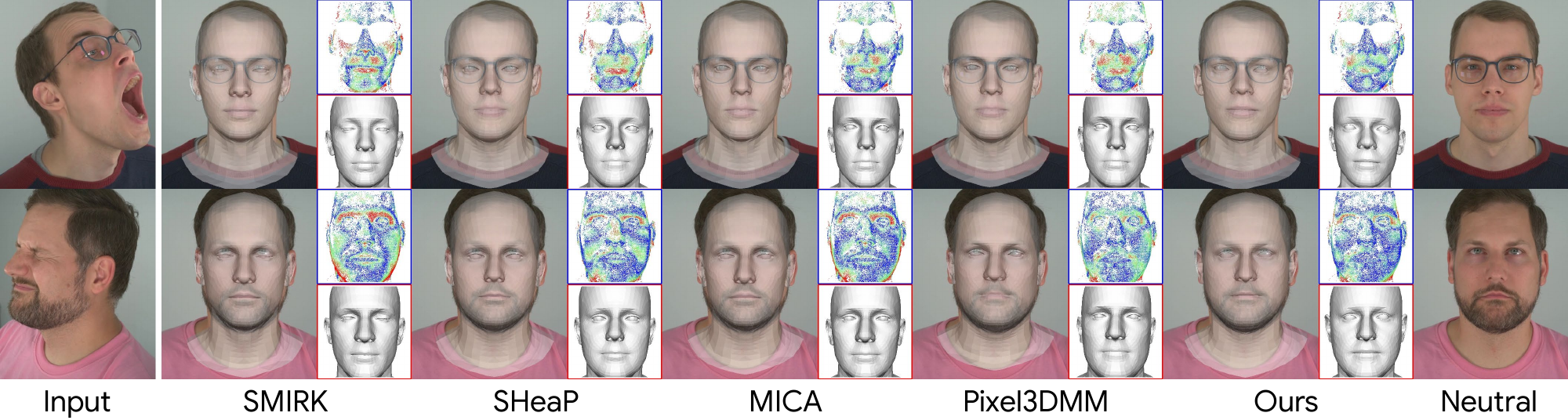}
    \caption{\textbf{Qualitative comparison on NeRSemble SVFR benchmark (neutral).} We show neutral-shape reconstructions and error maps from expressive monocular inputs, with the target neutral image shown in the last column. Our method produces cleaner identity geometry with fewer visible errors.}
    \label{fig:comp_nersemble_neutral}
\end{figure*}

\section{Experiments}

We evaluate our method from four aspects. First, we compare single-image fitting accuracy on the NeRSemble SVFR benchmarks. Second, we report results on the NoW benchmark to assess neutral-shape reconstruction. Third, we evaluate the quality and runtime performance of sequence reconstruction (both offline reconstruction and online tracking). Finally, we study the effects of the relative-depth prior, pose initialization, MICA initialization, and the Gauss-Newton solver through ablations.

\subsection{Implementation Details}
We summarize the dataset preprocessing and focal-length search used in our experiments here, and provide additional implementation details including network training and FLAME fitting hyperparameters in the supplementary material.

\noindent \textbf{Dataset preprocessing.} Following Pixel3DMM~\cite{Pixel3DMM}, we use three studio-captured datasets for prior network training: AVA-256~\cite{Ava256}, NPHM~\cite{NPHM}, and FaceScape~\cite{FaceScape}. For AVA-256, each frame is associated with a high-quality tracked mesh with consistent topology. We transfer the tracked mesh topology to FLAME and obtain high-quality FLAME registrations accordingly. Since AVA-256 is a video dataset, we further estimate SMIRK~\cite{SMIRK} expression coefficients and perform farthest point sampling for each subject to select around 60 frames with diverse expressions. After processing AVA-256, we train an initial prior network. For NPHM and FaceScape, although topology-consistent tracked meshes are not available, both datasets provide high-quality scan meshes. We therefore fit FLAME to these scans using Iterative Closest Point (ICP) together with the dense alignment predicted by the model pretrained on AVA-256, as illustrated in Fig.~\ref{fig:dataset}. In addition, NPHM does not provide the original captured multi-view images, but it provides textured scan meshes. We therefore render training images in Blender by randomly sampling HDRI environment lighting and camera viewpoints, as shown in Fig.~\ref{fig:dataset}. This preprocessing pipeline yields approximately 349K, 384K, and 244K training images from AVA-256, NPHM, and FaceScape, respectively.

\noindent \textbf{Focal-length search.} When camera intrinsics are not available, we estimate the focal length before fitting. For single-image fitting, the search is performed on the input image; for sequence reconstruction, it is performed on the first frame. We do not include the focal length directly in the Gauss-Newton variables. Instead, we perform a one-dimensional search outside the main fitting problem: for each candidate focal length, we construct the corresponding camera model, run a short fitting steps, and evaluate the resulting fitting energy. The focal length with the lowest energy is selected using golden-section search over a field-of-view range of \(5^\circ\) to \(40^\circ\) with \(5\) iterations.

\subsection{Baselines}
\noindent \textbf{Feed-forward methods.} We compare with representative feed-forward monocular face reconstruction methods, including DECA~\cite{DECA}, EMOCA~\cite{EMOCA}, TokenFace~\cite{TokenFace}, MICA~\cite{MICA}, SMIRK~\cite{SMIRK}, and SHeaP~\cite{SHeaP}. Among them, MICA estimates only the neutral facial shape. SMIRK and SHeaP are two recent strong baselines for expressive monocular reconstruction. SMIRK improves expressive reconstruction with an analysis-by-neural-synthesis training scheme, while SHeaP predicts head geometry under Gaussian-based rendering supervision.

\noindent \textbf{Optimization-based methods.} We also compare with optimization-based methods, including VHAP~\cite{VHAP}, FlowFace~\cite{FlowFace}, and Pixel3DMM~\cite{Pixel3DMM}. VHAP is a traditional sequence-based tracking method using photometric and landmark objectives, and we evaluate it only for offline sequence reconstruction and online tracking because the photometric constraints are not meaningful in the single-image fitting setting. FlowFace predicts dense UV-to-image correspondences for fitting, while Pixel3DMM predicts screen-space UV coordinates and surface normals for fitting.

\begin{table}[!t]
\centering
\scriptsize
\caption{\textbf{Results on NeRSemble SVFR benchmark (Official).} The official benchmark~\cite{NeRSemble,Pixel3DMM} evaluates 391 scans, with one view per scan. Our method achieves the best results on most metrics for both neutral and posed reconstruction.}
\begin{tabular}{lcccccc}
    \toprule
     & \multicolumn{3}{c}{Neutral} & \multicolumn{3}{c}{Posed} \\
    \cmidrule(lr){2-4} \cmidrule(lr){5-7}
    Method 
    & L1$\downarrow$ & L2$\downarrow$ & NC$\uparrow$
    & L1$\downarrow$ & L2$\downarrow$ & NC$\uparrow$ \\
    \midrule
    MICA~\cite{MICA}        
    & 1.680 & 1.136 & 0.8848 &   -   &   -   &   -    \\
    TokenFace~\cite{TokenFace}  
    &  -    &  -    &   -    & 2.627 & 1.779 & 0.8655 \\
    DECA~\cite{DECA}        
    & 2.078 & 1.402 & 0.8769 & 2.385 & 1.611 & 0.8710 \\
    EMOCA~\cite{EMOCA}       
    & 2.212 & 1.492 & 0.8741 & 2.636 & 1.777 & 0.8602 \\
    Skullptor~\cite{Skullptor}
    &   -   &   -   &   -    & 2.260 & 1.531 &   -    \\
    SMIRK~\cite{SMIRK}
    & 1.993 & 1.349 & 0.8820 & 2.276 & 1.533 & 0.8697 \\ 
    PartFusion
    & 2.022 & 1.363 & 0.8826 & 1.868 & 1.256 & 0.8828 \\ 
    PartFusionMI
    & 1.880 & 1.271 & 0.8852 & 1.700 & 1.145 & \textbf{0.8860} \\ 
    SHeaP~\cite{SHeaP}
    & 1.866 & 1.262 & 0.8818 & 2.083 & 1.407 & 0.8764 \\
    FlowFace~\cite{FlowFace}
    & 1.937 & 1.308 & 0.8804 & 1.973 & 1.330 & 0.8802 \\
    Pixel3DMM~\cite{Pixel3DMM}
    & \underline{1.654}  & \underline{1.119}  & \underline{0.8850} 
    & \underline{1.659}  & \underline{1.118}  & 0.8847 \\
    Ours    & \textbf{1.637} & \textbf{1.107} & \textbf{0.8857} 
            & \textbf{1.610} & \textbf{1.086} & \underline{0.8857} \\
    \bottomrule
\end{tabular}
\label{tab:svfr_official}
\end{table}

\begin{table}[!t]
\centering
\footnotesize
\caption{\textbf{Results on NoW Benchmark.} The NoW benchmark evaluates only neutral reconstruction. Our method outperforms the two most relevant dense-prior optimization baselines, FlowFace~\cite{FlowFace} and Pixel3DMM~\cite{Pixel3DMM}.}
\begin{tabular}{lccccc}
    \toprule
    Method & Median$\downarrow$ & Mean$\downarrow$ & Std$\downarrow$\\
    \midrule
    SHeaP~\cite{SHeaP}  & 0.95 & 1.18 & 0.99 \\
    MICA~\cite{MICA}    & 0.90 & 1.11 & 0.92 \\
    TokenFace~\cite{TokenFace}
    & \textbf{0.76} & \textbf{0.95} & \textbf{0.82} \\
    FlowFace~\cite{FlowFace}  & 0.87 & 1.07 & 0.88 \\
    Pixel3DMM~\cite{Pixel3DMM} & 0.87 & 1.07 & 0.89 \\
    Ours      & \underline{0.82} & \underline{1.02} & \underline{0.84} \\
    \bottomrule
\end{tabular}
\label{tab:now}
\end{table}

\subsection{Results on NeRSemble SVFR Benchmark}
\noindent \textbf{Official benchmark protocol.} The NeRSemble SVFR benchmark~\cite{NeRSemble,Pixel3DMM} evaluates single-image FLAME reconstruction on monocular views from the NeRSemble dataset~\cite{NeRSemble}. For each test scan, the benchmark reconstructs reference geometry from multi-view images using COLMAP and measures the distance from the reference point cloud to the submitted FLAME mesh after alignment. The official evaluation contains \(391\) scans and reports three metrics: L1 Chamfer distance, L2 Chamfer distance, and normal similarity between the predicted mesh and the reference point cloud. It evaluates two tasks: \emph{neutral reconstruction}, which measures the subject-specific neutral facial shape, and \emph{posed reconstruction}, which measures the expressive posed geometry of the input frame. Since the test annotations and evaluation code are not released, methods are evaluated by submitting reconstructed FLAME parameters to the official evaluation server. Table~\ref{tab:svfr_official} reports the official results.

\noindent \textbf{Official benchmark results.} Our method achieves the best overall performance on the official NeRSemble SVFR benchmark~\cite{NeRSemble} in both neutral and posed reconstruction. On neutral reconstruction, our method reduces the L1 error from \(1.654\) to \(1.637\) compared with Pixel3DMM~\cite{Pixel3DMM}, while also achieving the best L2 error and normal similarity. On posed reconstruction, our method similarly improves over Pixel3DMM, reducing the L1 error from \(1.659\) to \(1.610\) and the L2 error from \(1.118\) to \(1.086\). These results show that the proposed dense priors and Gauss-Newton solver improve reconstruction accuracy under the official evaluation protocol.

\noindent \textbf{Extended benchmark.} Because the official benchmark releases no ground-truth scans or evaluation code, we cannot inspect predictions qualitatively, run ablations, or analyze reconstruction behavior under controlled views. We therefore construct an extended NeRSemble-style benchmark of \(1{,}150\) scans from \(20\) subjects disjoint from the \(391\) official scans. Following Pixel3DMM~\cite{Pixel3DMM}, we reconstruct ground-truth pointclouds with COLMAP from the multi-view images. Our method achieves the best performance here as well. The detailed protocol and results are reported in the supplementary material.

\noindent \textbf{Qualitative comparison.} The qualitative comparisons in Fig.~\ref{fig:comp_nersemble} further illustrate the advantage of our method on challenging posed examples. Our reconstructions remain stable under large head rotations and strong expressions, while SMIRK~\cite{SMIRK} often fails to recover the correct expression and SHeaP~\cite{SHeaP} and Pixel3DMM~\cite{Pixel3DMM} tend to under-fit expressive regions. The difference is particularly visible in side-view mouth-opening cases, where dense 2D alignment alone is insufficient to resolve the full 3D facial structure. By combining UV-space correspondence with relative-depth constraints, our method provides stronger geometric priors to the fitting stage and better recovers the posed facial geometry. The neutral reconstruction comparisons in Fig.~\ref{fig:comp_nersemble_neutral} show a similar trend: our method produces stable identity geometry while still refining the initial identity estimate through optimization.

\noindent \textbf{Runtime.} On a single RTX 4090 GPU, the full single-image fitting pipeline takes \(64.4\,\mathrm{ms}\) with known camera intrinsics and \(181.9\,\mathrm{ms}\) when focal-length search is conducted. We provide a per-stage runtime breakdown in the supplementary material.

\subsection{Results on NoW Benchmark}
The NoW benchmark~\cite{NoW} evaluates only the neutral reconstruction task, but compared with the NeRSemble SVFR benchmark~\cite{NeRSemble,Pixel3DMM}, it covers a wider variety of identities, lighting conditions, hairstyles, head accessories, and other occlusions. As shown in Table~\ref{tab:now}, our method outperforms the two most related dense-prior optimization baselines, FlowFace~\cite{FlowFace} and Pixel3DMM~\cite{Pixel3DMM}. TokenFace~\cite{TokenFace} obtains the best NoW result, but it is not publicly available and performs poorly on the NeRSemble SVFR benchmark, especially for posed reconstruction. These results indicate that our method is also strong on the standard neutral-reconstruction benchmark, while its largest gains appear in challenging posed reconstruction and efficient optimization-based fitting.

\subsection{Sequence Reconstruction Results}
\noindent \textbf{Evaluation protocol.}
For sequence reconstruction, we evaluate monocular video-based neutral reconstruction on five NeRSemble sequences~\cite{NeRSemble} with noticeable head motion using a frontal camera. Since feed-forward monocular reconstruction methods such as MICA~\cite{MICA}, SMIRK~\cite{SMIRK}, and SHeaP~\cite{SHeaP} do not directly recover a single canonical identity from a video sequence, we evaluate them by averaging the predicted FLAME shape coefficients over all frames and then computing the neutral reconstruction metrics from the averaged shape. We also include VHAP in this sequence setting. This protocol allows us to compare how well different methods consolidate identity information from multiple frames of the same subject under changing poses.

\noindent \textbf{Offline reconstruction results.}
As shown in Table~\ref{tab:seq}, our offline sequence reconstruction achieves the best neutral reconstruction quality among all compared methods. Compared with the strongest previous monocular optimization baseline Pixel3DMM~\cite{Pixel3DMM}, our method reduces CD from \(1.76\) to \(1.69\), while also improving NC from \(0.930\) to \(0.933\) and \(R^{2.5}\) from \(76.9\) to \(78.5\). It also outperforms MICA~\cite{MICA}, which obtains CD \(1.85\), by a clear margin. Importantly, unlike our single-image fitting pipeline, this sequence experiment does not use MICA for identity initialization, but initializes the optimization from zero. This result indicates that with sufficiently rich priors across frames, the proposed fitting framework can recover accurate identity geometry directly from monocular video.

\begin{figure}[!t]
    \centering
    \includegraphics[width=0.7\linewidth]{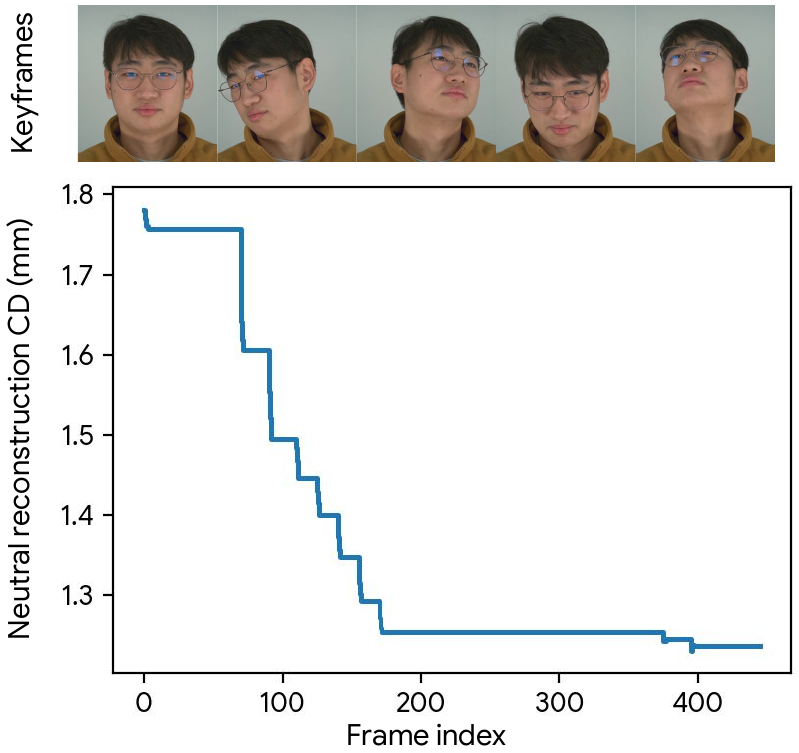}
    \caption{\textbf{Online reconstruction.} We visualize the online reconstruction process over time. The frames shown in the top row are the keyframes selected as the video proceeds, while the curve in the bottom row plots the neutral reconstruction error over time. As more informative keyframes are added, the online method progressively refines the shared identity and the neutral reconstruction error decreases.}
    \label{fig:online_recon}
\end{figure}

\noindent \textbf{Online tracking results.}
Our online tracking results remains close to the offline reconstruction while operating in the streaming setting. Each frame is processed only once in temporal order, and the online optimizer uses only the current frame and previously stored states, without access to future frames. Despite this stronger constraint, the online version achieves CD \(1.73\), NC \(0.930\), and \(R^{2.5}=77.8\), outperforming all previous baselines on CD and \(R^{2.5}\) and remaining close to the offline result. The qualitative behavior in Fig.~\ref{fig:online_recon} further shows that the neutral reconstruction error gradually decreases as more keyframes are selected and incorporated, demonstrating that the online tracking method can progressively refine identity geometry in a fully streaming setting.

\noindent \textbf{Runtime performance.}
The sequence runtime is measured on a \(25.5\)-second video containing \(1862\) frames at \(73\) FPS, using a single NVIDIA RTX 4090 GPU. As expected, feed-forward methods remain the fastest because they avoid iterative fitting, with SHeaP reaching \(343.96\) FPS in this setting. Among optimization-based methods, previous baselines are substantially slower: VHAP requires \(2058\) seconds and Pixel3DMM~\cite{Pixel3DMM} requires \(2994\) seconds on the evaluated sequences. In contrast, our offline sequence reconstruction finishes in \(84\) seconds while achieving better reconstruction accuracy. More importantly, the online version processes the same sequence in \(23\) seconds, corresponding to \(83.96\) FPS, and therefore runs faster than the input frame rate. A per-stage runtime breakdown is provided in the supplementary material.

\begin{table}[!t]
\footnotesize
\centering
\caption{\textbf{Comparison of quality and runtime performance of neutral reconstruction in monocular sequences.} We evaluate sequence-based neutral reconstruction on five NeRSemble videos~\cite{NeRSemble} with relatively large head motion. For the feed-forward methods above the horizontal line, we compute the final neutral-shape metrics from the average FLAME shape coefficients predicted over all frames. The optimization-based methods are listed below the line. Feed-forward methods remain the fastest, while our method achieves the best reconstruction quality and is substantially faster than previous optimization-based baselines. The online version remains close to the offline solver and runs in real time. The evaluation is conducted on videos with about 1.8K frames using a single RTX 4090 GPU.}
\begin{tabular}{lccc|cc}
    \toprule
    Method 
    & CD$\downarrow$ & NC$\uparrow$ & R$^{2.5}\uparrow$ & Runtime & FPS \\
    \midrule
    MICA      & 1.85 & 0.930 & 75.3 & \(18\,\mathrm{s}\) & 102.48 \\
    SMIRK     & 2.05 & 0.924 & 71.0 & \underline{\(17\,\mathrm{s}\)} & \underline{111.50} \\
    SHeaP     & 1.99 & 0.927 & 70.7 & \textbf{\(5\,\mathrm{s}\)} & \textbf{343.96} \\
    \midrule
    VHAP      & 2.21 & 0.924 & 66.7 & \(2058\,\mathrm{s}\) & - \\
    Pixel3DMM & 1.76 & 0.930 & 76.9 & \(2994\,\mathrm{s}\) & - \\
    Ours (offline) & \textbf{1.69} & \textbf{0.933} & \textbf{78.5} & \(84\,\mathrm{s}\) & - \\
    Ours (online)  & \underline{1.73} & \underline{0.930} & \underline{77.8} & \(23\,\mathrm{s}\) & 83.96 \\
    \bottomrule
\end{tabular}
\label{tab:seq}
\end{table}

\begin{table}[t]
\footnotesize
\centering
\caption{\textbf{Ablation studies on NeRSemble SVFR benchmark (Extended).} We study the effects of MICA~\cite{MICA} identity initialization, pose initialization, the relative-depth prior, and the fitting solver on the extended NeRSemble benchmark~\cite{NeRSemble,Pixel3DMM}. The full method achieves the best overall performance, showing that all four components contribute to the final reconstruction quality.}
\begin{tabular}{lcccccc}
    \toprule
     & \multicolumn{3}{c}{Neutral} & \multicolumn{3}{c}{Posed} \\
    \cmidrule(lr){2-4} \cmidrule(lr){5-7}
    Method 
    & CD$\downarrow$ & NC$\uparrow$ & R$^{2.5}\uparrow$
    & CD$\downarrow$ & NC$\uparrow$ & R$^{2.5}\uparrow$ \\
    \midrule
    w/o MICA       & 2.09 & \underline{0.926} & 69.2 & 1.60 & 0.921 & 80.4 \\
    w/o pose init  & \underline{1.92} & 0.925 & \underline{73.4} & \underline{1.59} & 0.921 & \underline{80.8} \\
    w/o depth      & 1.93 & 0.923 & 72.5 & 1.64 & \underline{0.922} & 79.5 \\
    Adam optimizer & 2.04 & \underline{0.926} & 70.9 & 1.63 & \textbf{0.924} & 79.7 \\
    \midrule
    Ours    & \textbf{1.86} & \textbf{0.927} & \textbf{74.6} 
            & \textbf{1.56} & \underline{0.922} & \textbf{81.4} \\
    \bottomrule
\end{tabular}
\label{tab:abla}
\end{table}

\begin{figure}[!t]
    \centering
    \includegraphics[width=1.0\linewidth]{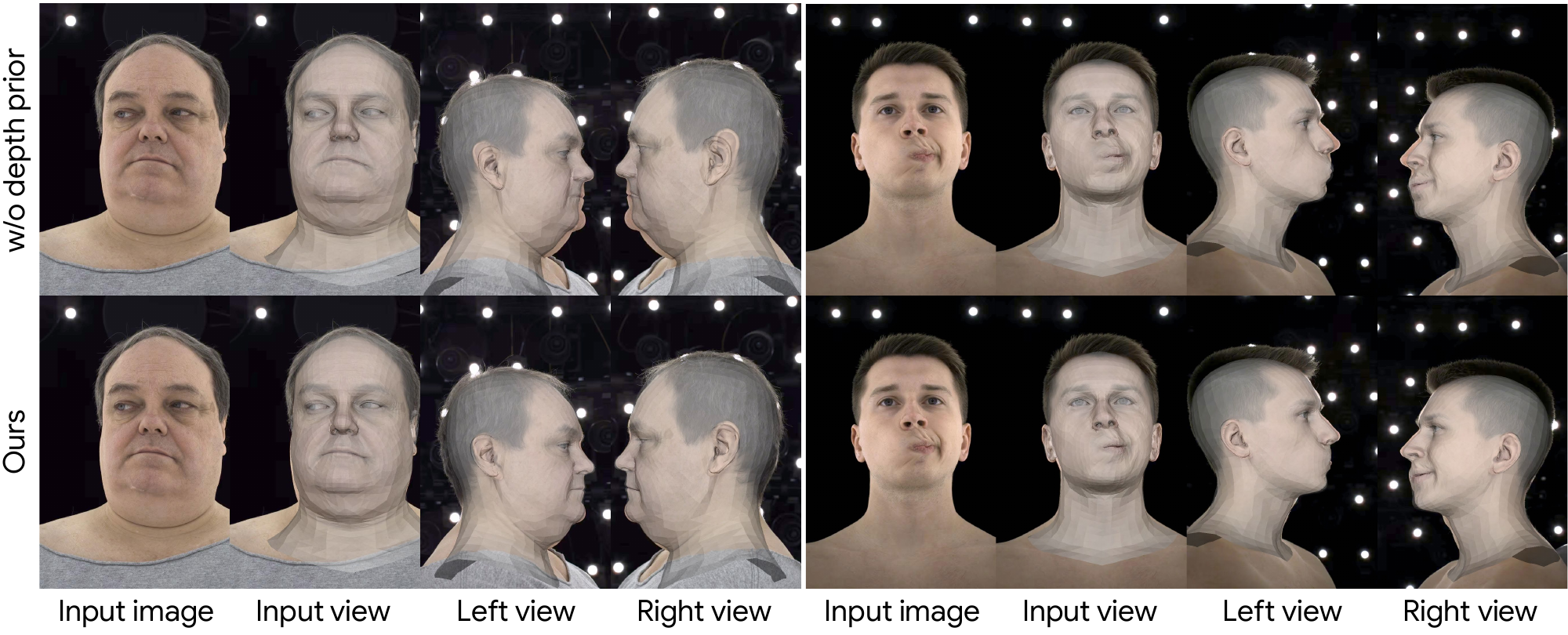}
    \caption{\textbf{Ablation of relative-depth prior.} In the input view, both variants can align the reconstructed mesh to the image reasonably well. However, under side views, using correspondence alone provides only 2D alignment without sufficient 3D geometric constraints, which leads to clear reconstruction errors in regions such as the chin (left) and the nose (right). Adding the relative-depth prior resolves these ambiguities and produces more accurate 3D facial geometry.}
    \label{fig:abla_depth}
\end{figure}

\begin{figure}[!t]
    \centering
    \includegraphics[width=1.0\linewidth]{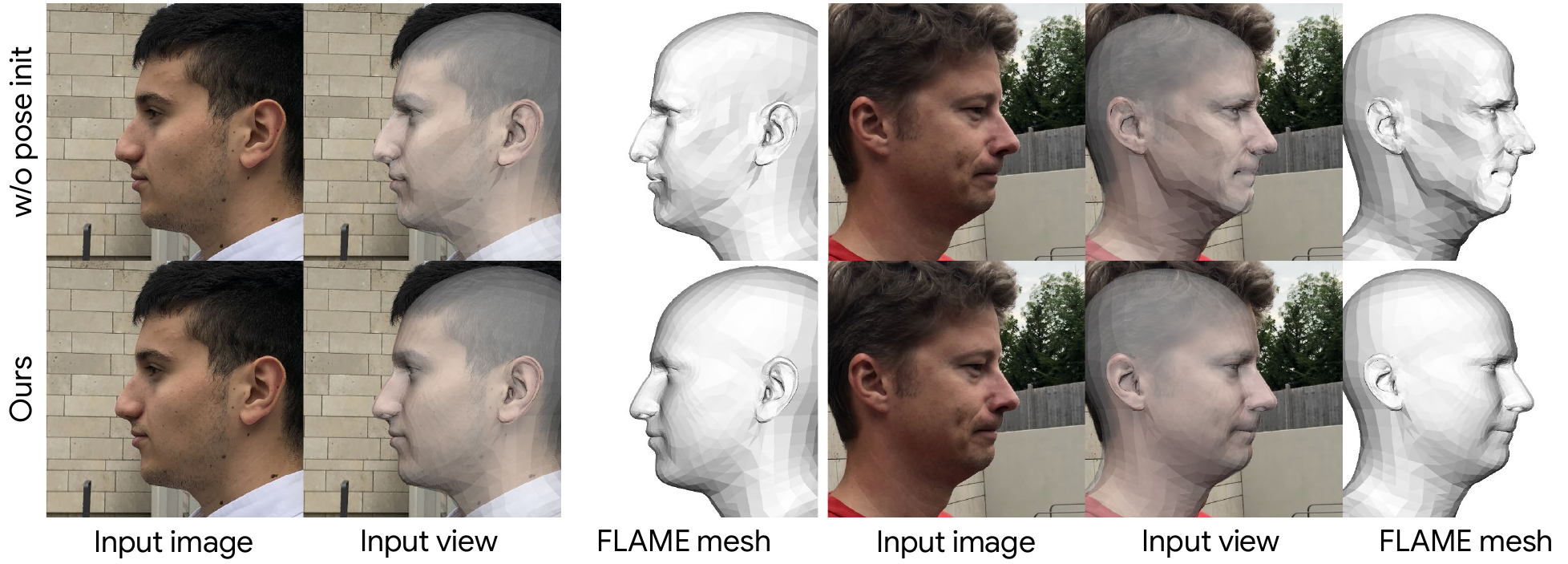}
    \caption{\textbf{Ablation of pose-initialization stage.} Under large side-view inputs, directly optimizing pose and expression together often traps the solver in poor local minima, so the dynamic-parameter update compensates for incorrect pose and causes severe mesh distortion. A short pose-only initialization provides a better starting point and makes the subsequent full optimization much more stable.}
    \label{fig:abla_pose_init}
\end{figure}

\subsection{Ablation Studies}
\noindent \textbf{Impact of dense priors.} We analyze the contribution of the relative-depth prior by removing the depth term and optimizing the model using correspondence alone. As shown in Fig.~\ref{fig:abla_depth}, dense 2D alignment already provides strong UV-to-image constraints, but remains ambiguous in depth. This is particularly evident under side views, where the depth of the chin and nose cannot be fully resolved from alignment cues alone. By introducing the relative-depth prior, our method receives additional geometric constraints along the viewing direction, which helps recover more plausible facial structure in these ambiguous cases. This effect is also reflected in the posed reconstruction results in Table~\ref{tab:abla}, where removing the depth prior leads to worse geometric accuracy.

\noindent \textbf{Impact of pose initialization.} We evaluate the effect of the pose initialization stage by disabling the pose-only optimization before the full fitting procedure. Because pose estimation is highly non-linear, directly optimizing pose and expression from a generic initialization can trap the solver in poor local minima, especially for large head rotations. As illustrated in Fig.~\ref{fig:abla_pose_init}, without pose initialization the reconstructed mesh can become severely distorted under profile or near-profile views. A short pose-only initialization stage provides a better starting point for subsequent optimization and improves robustness in challenging poses. This behavior is also confirmed quantitatively in Table~\ref{tab:abla}, where removing pose initialization degrades reconstruction performance.

\noindent \textbf{Impact of MICA.} We also study the role of MICA-based identity initialization. MICA~\cite{MICA} provides a strong subject-specific identity prior, which helps disentangle the identity and expression spaces of FLAME before iterative fitting. In practice, this leads to a cleaner identity estimate and reduces the tendency of expression-dependent geometry to leak into the identity component. As a result, the neutral facial shape can be reconstructed more accurately when MICA is used for initialization. This trend is supported by the quantitative results in Table~\ref{tab:abla}, where removing MICA causes a clear drop on the neutral reconstruction task.

\begin{figure}
    \centering
    \includegraphics[width=0.8\linewidth]{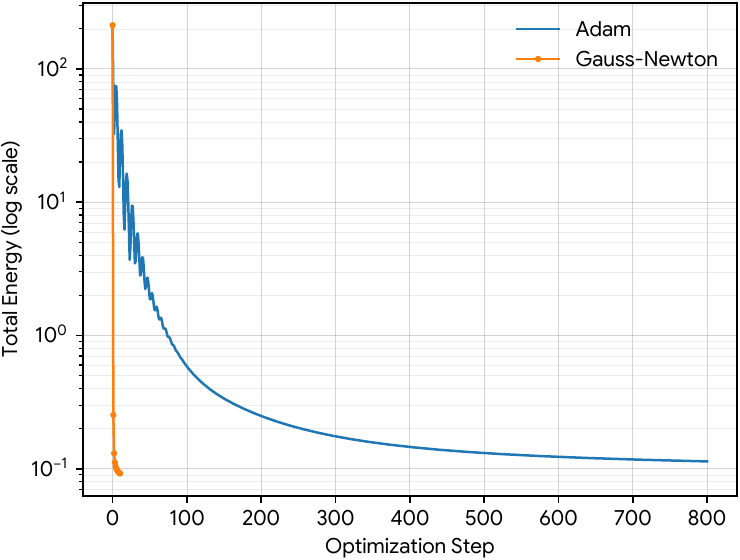}
    \caption{\textbf{Adam vs. Gauss-Newton fitting optimization.} We compare Adam optimization with our Gauss-Newton solver for single-image FLAME fitting. Our solver decreases the fitting energy faster and reaches a lower final energy with far fewer iterations.}
    \label{fig:solver_comp}
\end{figure}

\noindent \textbf{Impact of the Gauss-Newton solver.} We further evaluate the impact of our Gauss-Newton optimizer by replacing it with Adam. As shown in Fig.~\ref{fig:solver_comp}, our solver reduces the fitting energy much more rapidly and converges to a lower final energy within far fewer iterations. This efficiency gap is substantial: Table~\ref{tab:gn_adam_compare} shows that Adam requires 800 steps and 18 seconds to optimize a single image, whereas our Gauss-Newton solver reaches a lower final energy with only 10 steps and \(29.64\,\mathrm{ms}\). Therefore, the proposed solver improves optimization speed by several orders of magnitude, making real-time fitting practical. In terms of reconstruction quality, the NeRSemble results (Table~\ref{tab:abla}) also favor our solver overall, indicating that this large gain in efficiency does not come at the cost of accuracy.

\begin{table}[]
    \centering
    \footnotesize
    \caption{\textbf{Adam vs. Gauss-Newton fitting optimization.} We report the runtime and final fitting energy for the same single-image FLAME fitting comparison as in Fig.~\ref{fig:solver_comp}. Our Gauss-Newton solver reaches a slightly lower final fitting energy than Adam, while reducing the per-step runtime by about \(8\times\), the number of optimization steps by about \(80\times\), and the total optimization time by about \(600\times\).}
    \label{tab:gn_adam_compare}
    \begin{tabular}{lcc}
        \toprule
        Method & Adam & Gauss-Newton \\
        \midrule
        Total time      & \(18\,\mathrm{s}\) & \(29.64\,\mathrm{ms}\) \\
        Total steps     & 800 & 10  \\
        Per step time   & \(23\,\mathrm{ms}\) & \(2.96\,\mathrm{ms}\) \\
        Final energy    & 0.1143 & 0.0901 \\
        \bottomrule
    \end{tabular}
\end{table}

\subsection{In-the-wild Results}
In Fig.~\ref{fig:in_the_wild}, we provide qualitative examples on challenging in-the-wild images from the VFHQ dataset~\cite{VFHQ}. The examples cover large appearance variations, including diverse skin tones, complex lighting and shadow effects, common occlusions such as eyeglasses, hair, and facial hair, as well as large head poses. Despite these challenges, our method still produces stable alignments and plausible FLAME reconstructions, suggesting that the predicted dense priors can generalize beyond the controlled training distribution. Additional in-the-wild qualitative comparisons with representative baselines are provided in Fig.~\ref{fig:in_the_wild_compare} of the supplementary material.

\section{Limitations}
Our method is still bounded by the expressiveness of the FLAME model. Although the proposed dense priors provide strong constraints for identity, expression, pose, and translation fitting, the final reconstruction remains a parametric FLAME mesh. Therefore, it cannot explicitly recover geometry outside the FLAME topology, such as hair, teeth, tongue, clothing, accessories, or fine-scale skin details. This limitation is shared by FLAME-based reconstruction methods and suggests that combining our efficient Gauss-Newton solver with richer head or appearance models would be an important future direction.

The current system is evaluated primarily in a single-person monocular setting on a high-end GPU. The online tracker runs in real time on an RTX 4090, but its speed and stability on lower-power devices, multi-person videos, or long unconstrained streams remain to be studied. In addition, when camera intrinsics are unknown, the optional focal-length search adds noticeable cost to single-image fitting. Future work could further reduce this overhead and extend the framework to broader deployment settings.

\section{Conclusion}
We presented a real-time optimization-based framework for monocular 3D face reconstruction from learned dense priors. The central idea is to make dense priors compatible with fast nonlinear least-squares fitting: UV-space correspondence provides vertex-wise UV-to-image alignment, while relative depth supplies geometric constraints along the viewing direction. From these priors, FLAME fitting becomes a compact nonlinear least-squares problem solved by a Gauss-Newton solver with analytic Jacobians and custom CUDA kernels. Experiments show that the proposed design improves accuracy over recent dense-prior optimization baselines on NeRSemble SVFR benchmark and also outperforms FlowFace and Pixel3DMM on NoW. The same solver further supports offline sequence reconstruction and online tracking, demonstrating that high-quality optimization-based face reconstruction can be made practical for real-time monocular video.

\section{Acknowledgements}
This work is partially supported by NSF China (No. 62421003) and the XPLORER PRIZE.
{
    \small
    \bibliographystyle{ieeenat_fullname}
    \bibliography{main}

@String(CVPR= {IEEE Conf. Comput. Vis. Pattern Recog.})

@String(ICCV= {Int. Conf. Comput. Vis.})

@String(ECCV= {Eur. Conf. Comput. Vis.})

@String(CVPRW= {IEEE Conf. Comput. Vis. Pattern Recog. Worksh.})

@String(CVPR  = {CVPR})

@String(ICCV  = {ICCV})

@String(ECCV  = {ECCV})

@String(CVPRW= {CVPRW})

@article{CaoTOG13,
  title = {{3D} Shape Regression for Real-Time Facial Animation},
  author = {Cao, Chen and Weng, Yanlin and Lin, Stephen and Zhou, Kun},
  journal = {ACM Trans. Graph.},
  year = {2013},
  volume = {32},
  number = {4},
  doi = {10.1145/2461912.2462012},
  issn = {0730-0301},
  month = {jul},
}

@article{DDE,
  title = {Displaced Dynamic Expression Regression for Real-Time Facial Tracking and Animation},
  author = {Cao, Chen and Hou, Qiming and Zhou, Kun},
  journal = {ACM Trans. Graph.},
  year = {2014},
  volume = {33},
  number = {4},
  doi = {10.1145/2601097.2601204},
  issn = {0730-0301},
  month = {jul},
}

@article{CaoTOG15,
  title = {Real-Time High-Fidelity Facial Performance Capture},
  author = {Cao, Chen and Bradley, Derek and Zhou, Kun and Beeler, Thabo},
  journal = {ACM Trans. Graph.},
  year = {2015},
  volume = {34},
  number = {4},
  doi = {10.1145/2766943},
  issn = {0730-0301},
  month = {jul},
}

@article{LiHaoCorrectives13,
  title = {Realtime Facial Animation with On-the-Fly Correctives},
  author = {Li, Hao and Yu, Jihun and Ye, Yuting and Bregler, Chris},
  journal = {ACM Trans. Graph.},
  year = {2013},
  volume = {32},
  number = {4},
  doi = {10.1145/2461912.2462019},
  issn = {0730-0301},
  month = {jul},
}

@article{BouazizOnline13,
  title = {Online Modeling for Realtime Facial Animation},
  author = {Bouaziz, Sofien and Wang, Yangang and Pauly, Mark},
  journal = {ACM Trans. Graph.},
  year = {2013},
  volume = {32},
  number = {4},
  doi = {10.1145/2461912.2461976},
  issn = {0730-0301},
  month = {jul},
}

@article{WeisePerformance11,
  title = {Realtime Performance-Based Facial Animation},
  author = {Weise, Thibaut and Bouaziz, Sofien and Li, Hao and Pauly, Mark},
  journal = {ACM Trans. Graph.},
  year = {2011},
  volume = {30},
  number = {4},
  doi = {10.1145/2010324.1964972},
  issn = {0730-0301},
  month = {jul},
}

@inproceedings{SMIRK,
  title = {{3D} Facial Expressions through Analysis-by-Neural-Synthesis},
  author = {Retsinas, George and Filntisis, Panagiotis P. and Danecek, Radek and Abrevaya, Victoria F. and Roussos, Anastasios and Bolkart, Timo and Maragos, Petros},
  booktitle = {2024 IEEE/CVF Conference on Computer Vision and Pattern Recognition (CVPR)},
  year = {2024},
  doi = {10.1109/cvpr52733.2024.00241},
  month = {apr},
}

@article{ThiesReenactment15,
  title = {Real-Time Expression Transfer for Facial Reenactment},
  author = {Thies, Justus and Zollh{\"o}fer, Michael and Nie{\ss}ner, Matthias and Valgaerts, Levi and Stamminger, Marc and Theobalt, Christian},
  journal = {ACM Trans. Graph.},
  year = {2015},
  volume = {34},
  number = {6},
  doi = {10.1145/2816795.2818056},
  issn = {0730-0301},
  month = {nov},
}

@article{LookMa,
  title = {Look Ma, No Markers: Holistic Performance Capture without the Hassle},
  author = {Hewitt, Charlie and Saleh, Fatemeh and Aliakbarian, Sadegh and Petikam, Lohit and Rezaeifar, Shideh and Florentin, Louis and Hosenie, Zafiirah and Cashman, Thomas J. and Valentin, Julien and Cosker, Darren and Baltrusaitis, Tadas},
  journal = {ACM Trans. Graph.},
  year = {2024},
  volume = {43},
  number = {6},
  doi = {10.1145/3687772},
  issn = {0730-0301},
  month = {nov},
}

@article{CaoStabilized18,
  title = {Stabilized Real-Time Face Tracking via a Learned Dynamic Rigidity Prior},
  author = {Cao, Chen and Chai, Menglei and Woodford, Oliver and Luo, Linjie},
  journal = {ACM Trans. Graph.},
  year = {2018},
  volume = {37},
  number = {6},
  doi = {10.1145/3272127.3275093},
  issn = {0730-0301},
  month = {dec},
}

@misc{WoodDense22,
  title = {{3D} Face Reconstruction with Dense Landmarks},
  author = {Wood, Erroll and Baltrusaitis, Tadas and Hewitt, Charlie and Johnson, Matthew and Shen, Jingjing and Milosavljevic, Nikola and Wilde, Daniel and Garbin, Stephan and Raman, Chirag and Shotton, Jamie and Sharp, Toby and Stojiljkovic, Ivan and Cashman, Tom and Valentin, Julien},
  year = {2022},
  month = {apr},
}

@article{FaceVR,
  title = {{FaceVR}: Real-Time Gaze-Aware Facial Reenactment in Virtual Reality},
  author = {Thies, Justus and Zollh{\"o}fer, Michael and Stamminger, Marc and Theobalt, Christian and Nie{\ss}ner, Matthias},
  journal = {ACM Trans. Graph.},
  year = {2018},
  volume = {37},
  number = {2},
  doi = {10.1145/3182644},
  issn = {0730-0301},
  month = {jun},
}

@inproceedings{MICA,
  title = {Towards Metrical Reconstruction Of Human Faces},
  author = {Zielonka, Wojciech and Bolkart, Timo and Thies, Justus},
  booktitle = {Computer {Vision} -- {ECCV} 2022: 17th {European Conference}, {Tel Aviv}, {Israel}, {October} 23--27, 2022, {Proceedings}, {Part XIII}},
  year = {2022},
  pages = {250--269},
  publisher = {Springer-Verlag},
  doi = {10.1007/978-3-031-19778-9_15},
  address = {Berlin, Heidelberg},
  isbn = {978-3-031-19777-2},
}

@article{Face2Face,
  title = {{Face2Face}: Real-Time Face Capture and Reenactment of {RGB} Videos},
  author = {Thies, Justus and Zollh{\"o}fer, Michael and Stamminger, Marc and Theobalt, Christian and Nie{\ss}ner, Matthias},
  journal = {Commun. ACM},
  year = {2018},
  volume = {62},
  number = {1},
  pages = {96--104},
  doi = {10.1145/3292039},
  issn = {0001-0782},
  month = {dec},
}

@inproceedings{FlowFace,
  title = {{3D} Face Tracking from {2D} Video through Iterative Dense {UV} to Image Flow},
  author = {Taubner, Felix and Raina, Prashant and Tuli, Mathieu and Teh, Eu Wern and Lee, Chul and Huang, Jinmiao},
  booktitle = {2024 IEEE/CVF Conference on Computer Vision and Pattern Recognition (CVPR)},
  year = {2024},
  doi = {10.1109/cvpr52733.2024.00123},
  month = {apr},
}

@inproceedings{SHeaP,
  title = {{SHeaP}: Self-Supervised Head Geometry Predictor Learned via {2D} Gaussians},
  author = {Schoneveld, Liam and Chen, Zhe and Davoli, Davide and Tang, Jiapeng and Terazawa, Saimon and Nishino, Ko and Nie{\ss}ner, Matthias},
  booktitle = {Proceedings of the IEEE/CVF International Conference on Computer Vision (ICCV)},
  year = {2025},
  month = {apr},
}

@inproceedings{Pixel3DMM,
  title={Pixel3{DMM}: Versatile Screen-Space Priors for Single-Image 3D Face Reconstruction},
  author = {Giebenhain, Simon and Kirschstein, Tobias and R{\"u}nz, Martin and Agapito, Lourdes and Nie{\ss}ner, Matthias},
  booktitle={The Fourteenth International Conference on Learning Representations},
  year={2026},
  url={https://openreview.net/forum?id=UmOdd5KQ8K}
}

@inproceedings{TokenFace,
  title = {Accurate {3D} Face Reconstruction with Facial Component Tokens},
  author = {Zhang, Tianke and Chu, Xuangeng and Liu, Yunfei and Lin, Lijian and Yang, Zhendong and Xu, Zhengzhuo and Cao, Chengkun and Yu, Fei and Zhou, Changyin and Yuan, Chun and Li, Yu},
  booktitle = {2023 {IEEE}/{CVF International Conference} on {Computer Vision} ({ICCV})},
  year = {2023},
  pages = {8999--9008},
  doi = {10.1109/ICCV51070.2023.00829},
}

@article{DECA,
  title = {Learning an Animatable Detailed {3D} Face Model from In-The-Wild Images},
  author = {Feng, Yao and Feng, Haiwen and Black, Michael J. and Bolkart, Timo},
  journal = {ACM Transactions on Graphics},
  year = {2021},
  doi = {10.1145/3450626.3459936},
  month = {jun},
}

@misc{Pix2NPHM,
  title = {{Pix2NPHM}: Learning to Regress {NPHM} Reconstructions From a Single Image},
  author = {Giebenhain, Simon and Kirschstein, Tobias and Schoneveld, Liam and Davoli, Davide and Chen, Zhe and Nie{\ss}ner, Matthias},
  year = {2025},
  doi = {10.48550/arXiv.2512.17773},
  archiveprefix = {arXiv},
  eprint = {2512.17773},
  month = {dec},
}

@inproceedings{Ava256,
  title = {Codec Avatar Studio: Paired Human Captures for Complete, Driveable, and Generalizable Avatars},
  author = {Martinez, Julieta and Kim, Emily and Romero, Javier and Bagautdinov, Timur and Saito, Shunsuke and Yu, Shoou-I and Anderson, Stuart and Zollh{\"o}fer, Michael and Wang, Te-Li and Bai, Shaojie and Li, Chenghui and Wei, Shih-En and Joshi, Rohan and Borsos, Wyatt and Simon, Tomas and Saragih, Jason and Theodosis, Paul and Greene, Alexander and Josyula, Anjani and Maeta, Silvio Mano and Jewett, Andrew I. and Venshtain, Simon and Heilman, Christopher and Chen, Yueh-Tung and Fu, Sidi and Elshaer, Mohamed Ezzeldin A. and Du, Tingfang and Wu, Longhua and Chen, Shen-Chi and Kang, Kai and Wu, Michael and Emad, Youssef and Longay, Steven and Brewer, Ashley and Shah, Hitesh and Booth, James and Koska, Taylor and Haidle, Kayla and Andromalos, Matt and Hsu, Joanna and Dauer, Thomas and Selednik, Peter and Godisart, Tim and Ardisson, Scott and Cipperly, Matthew and Humberston, Ben and Farr, Lon and Hansen, Bob and Guo, Peihong and Braun, Dave and Krenn, Steven and Wen, He and Evans, Lucas and Fadeeva, Natalia and Stewart, Matthew and Schwartz, Gabriel and Gupta, Divam and Moon, Gyeongsik and Guo, Kaiwen and Dong, Yuan and Xu, Yichen and Shiratori, Takaaki and Prada, Fabian and Pires, Bernardo R. and Peng, Bo and Buffalini, Julia and Trimble, Autumn and McPhail, Kevyn and Schoeller, Melissa and Sheikh, Yaser},
  booktitle = {Advances in {Neural Information Processing Systems}},
  year = {2024},
  volume = {37},
  pages = {83008--83023},
  publisher = {Curran Associates, Inc.},
  editor = {Globerson, A. and Mackey, L. and Belgrave, D. and Fan, A. and Paquet, U. and Tomczak, J. and Zhang, C.},
}

@article{FaceScape,
  title = {{FaceScape}: {3D} Facial Dataset and Benchmark for Single-View {3D} Face Reconstruction},
  author = {Zhu, Hao and Yang, Haotian and Guo, Longwei and Zhang, Yidi and Wang, Yanru and Huang, Mingkai and Wu, Menghua and Shen, Qiu and Yang, Ruigang and Cao, Xun},
  journal = {IEEE Transactions on Pattern Analysis and Machine Intelligence},
  year = {2023},
  doi = {10.1109/tpami.2023.3307338},
  month = {sep},
}

@article{NeRSemble,
  title = {{NeRSemble}: Multi-view Radiance Field Reconstruction of Human Heads},
  author = {Kirschstein, Tobias and Qian, Shenhan and Giebenhain, Simon and Walter, Tim and Nie{\ss}ner, Matthias},
  journal = {ACM Transactions on Graphics},
  year = {2023},
  doi = {10.1145/3592455},
  month = {may},
}

@article{FLAME,
    author = {Li, Tianye and Bolkart, Timo and Black, Michael J. and Li, Hao and Romero, Javier},
    title = {Learning a model of facial shape and expression from 4D scans},
    year = {2017},
    issue_date = {December 2017},
    publisher = {Association for Computing Machinery},
    address = {New York, NY, USA},
    volume = {36},
    number = {6},
    issn = {0730-0301},
    url = {https://doi.org/10.1145/3130800.3130813},
    doi = {10.1145/3130800.3130813},
    journal = {ACM Trans. Graph.},
    month = nov,
    articleno = {194},
    numpages = {17},
}

@InProceedings{RGBAvatar,
    author    = {Li, Linzhou and Li, Yumeng and Weng, Yanlin and Zheng, Youyi and Zhou, Kun},
    title     = {RGBAvatar: Reduced Gaussian Blendshapes for Online Modeling of Head Avatars},
    booktitle = {Proceedings of the IEEE/CVF Conference on Computer Vision and Pattern Recognition (CVPR)},
    month     = {June},
    year      = {2025},
    pages     = {10747-10757}
}

@inproceedings{3DGB,
  author     = {Shengjie Ma and Yanlin Weng and Tianjia Shao and Kun Zhou},
  title      = {3D Gaussian Blendshapes for Head Avatar Animation},
  booktitle  = {ACM SIGGRAPH Conference Proceedings, Denver, CO, United States, July 28 - August 1, 2024},
  year       = {2024},
}

@inproceedings{MAMMA,
  title     = {{MAMMA}: {Markerless Accurate Multi-person Motion Acquisition}},
  author    = {Cuevas Velasquez, Hanz and Yiannakidis, Anastasios and Shin, Soyong and Becherini, Giorgio and H{\"o}schle, Markus and Tesch, Joachim and Obersat, Taylor and Alexiadis, Tsvetelina and Halilaj, Eni and Black, Michael J.},
  booktitle = {Proceedings of the IEEE/CVF Conference on Computer Vision and Pattern Recognition (CVPR)},
  year      = {2026}
}

@ARTICLE{FaceWarehouse,
  author={Cao, Chen and Weng, Yanlin and Zhou, Shun and Tong, Yiying and Zhou, Kun},
  journal={IEEE Transactions on Visualization and Computer Graphics}, 
  title={{FaceWarehouse}: A 3D Facial Expression Database for Visual Computing}, 
  year={2014},
  volume={20},
  number={3},
  pages={413-425},
  doi={10.1109/TVCG.2013.249}
}

@misc{VHAP,
  title={{VHAP}: Versatile Head Alignment with Adaptive Appearance Priors},
  author={Qian, Shenhan},
  year={2024},
  month={sep},
  doi={10.5281/zenodo.14988309},
  url={https://github.com/ShenhanQian/VHAP}
}

@misc{MetricalTracker,
  title={Metrical Monocular Photometric Tracker},
  author = {Zielonka, Wojciech and Bolkart, Timo and Thies, Justus},
  year={2022},
  url={https://github.com/Zielon/metrical-tracker}
}

@inproceedings{EMOCA,
  title = {{EMOCA}: {E}motion Driven Monocular Face Capture and Animation},
  author = {Danecek, Radek and Black, Michael J. and Bolkart, Timo},
  booktitle = {Conference on Computer Vision and Pattern Recognition (CVPR)},
  pages = {20311--20322},
  year = {2022}
}

@inproceedings{GNLL,
     author = {Kendall, Alex and Gal, Yarin},
     booktitle = {Advances in Neural Information Processing Systems},
     editor = {I. Guyon and U. Von Luxburg and S. Bengio and H. Wallach and R. Fergus and S. Vishwanathan and R. Garnett},
     pages = {},
     publisher = {Curran Associates, Inc.},
     title = {What Uncertainties Do We Need in Bayesian Deep Learning for Computer Vision?},
     volume = {30},
     year = {2017}
}

@InProceedings{NPHM,
    author    = {Giebenhain, Simon and Kirschstein, Tobias and Georgopoulos, Markos and R\"unz, Martin and Agapito, Lourdes and Nie{\ss}ner, Matthias},
    title     = {Learning Neural Parametric Head Models},
    booktitle = {Proceedings of the IEEE/CVF Conference on Computer Vision and Pattern Recognition (CVPR)},
    month     = {June},
    year      = {2023},
    pages     = {21003-21012}
}

@inproceedings{NoW,
    title = {Learning to Regress {3D} Face Shape and Expression from an Image without {3D} Supervision},
    author = {Sanyal, Soubhik and Bolkart, Timo and Feng, Haiwen and Black, Michael},
    booktitle = {Proceedings IEEE Conf. on Computer Vision and Pattern Recognition (CVPR)},
    month = jun,
    pages = {7763--7772},
    year = {2019},
    month_numeric = {6} 
}

@InProceedings{VFHQ,
    author = {Liangbin Xie and Xintao Wang and Honglun Zhang and Chao Dong and Ying Shan},
    title = {{VFHQ}: A High-Quality Dataset and Benchmark for Video Face Super-Resolution},
    booktitle={The IEEE Conference on Computer Vision and Pattern Recognition Workshops (CVPRW)},
    year = {2022}
}

@inproceedings{BlanzVetter99,
  title = {A Morphable Model for the Synthesis of {3D} Faces},
  author = {Blanz, Volker and Vetter, Thomas},
  booktitle = {Proceedings of the 26th Annual Conference on Computer Graphics and Interactive Techniques},
  pages = {187--194},
  year = {1999},
  doi = {10.1145/311535.311556}
}

@inproceedings{BFM09,
  title = {A {3D} Face Model for Pose and Illumination Invariant Face Recognition},
  author = {Paysan, Pascal and Knothe, Reinhard and Amberg, Brian and Romdhani, Sami and Vetter, Thomas},
  booktitle = {2009 Sixth IEEE International Conference on Advanced Video and Signal Based Surveillance},
  pages = {296--301},
  year = {2009},
  doi = {10.1109/AVSS.2009.58}
}

@inproceedings{Deep3DFaceRecon19,
  title = {Accurate {3D} Face Reconstruction with Weakly-Supervised Learning: From Single Image to Image Set},
  author = {Deng, Yu and Yang, Jiaolong and Xu, Sicheng and Chen, Dong and Jia, Yunde and Tong, Xin},
  booktitle = {Proceedings of the IEEE/CVF Conference on Computer Vision and Pattern Recognition Workshops (CVPRW)},
  year = {2019}
}

@inproceedings{ThreeDDFAv2,
  title = {Towards Fast, Accurate and Stable {3D} Dense Face Alignment},
  author = {Guo, Jianzhu and Zhu, Xiangyu and Yang, Yang and Yang, Fan and Lei, Zhen and Li, Stan Z.},
  booktitle = {Computer Vision -- ECCV 2020},
  pages = {152--168},
  year = {2020}
}

@inproceedings{HRN,
  title = {A Hierarchical Representation Network for Accurate and Detailed Face Reconstruction from In-the-Wild Images},
  author = {Lei, Biwen and Ren, Jianqiang and Feng, Mengyang and Cui, Miaomiao and Xie, Xuansong},
  booktitle = {Proceedings of the IEEE/CVF Conference on Computer Vision and Pattern Recognition (CVPR)},
  pages = {394--403},
  year = {2023}
}

@article{SADRNet,
  title = {{SADRNet}: Self-Aligned Dual Face Regression Networks for Robust {3D} Dense Face Alignment and Reconstruction},
  author = {Ruan, Zeyu and Zou, Changqing and Wu, Longhai and Wu, Gangshan and Wang, Limin},
  journal = {IEEE Transactions on Image Processing},
  volume = {30},
  pages = {5793--5806},
  year = {2021},
  doi = {10.1109/TIP.2021.3087397}
}

@inproceedings{GaussianAvatars,
  title = {{GaussianAvatars}: Photorealistic Head Avatars with Rigged {3D} Gaussians},
  author = {Qian, Shenhan and Kirschstein, Tobias and Schoneveld, Liam and Davoli, Davide and Giebenhain, Simon and Nie{\ss}ner, Matthias},
  booktitle = {Proceedings of the IEEE/CVF Conference on Computer Vision and Pattern Recognition (CVPR)},
  pages = {20299--20309},
  year = {2024}
}

@inproceedings{RichardsonSynthetic16,
  title = {{3D} Face Reconstruction by Learning from Synthetic Data},
  author = {Richardson, Elad and Sela, Matan and Kimmel, Ron},
  booktitle = {2016 Fourth International Conference on 3D Vision (3DV)},
  pages = {460--467},
  year = {2016},
  doi = {10.1109/3DV.2016.56}
}

@inproceedings{RichardsonDetailed17,
  title = {Learning Detailed Face Reconstruction from a Single Image},
  author = {Richardson, Elad and Sela, Matan and Or-El, Roy and Kimmel, Ron},
  booktitle = {Proceedings of the IEEE Conference on Computer Vision and Pattern Recognition (CVPR)},
  pages = {1259--1268},
  year = {2017}
}

@inproceedings{Tran3DMM17,
  title = {Regressing Robust and Discriminative {3D} Morphable Models with a Very Deep Neural Network},
  author = {Tran, Anh Tuan and Hassner, Tal and Masi, Iacopo and Medioni, G{\'e}rard},
  booktitle = {Proceedings of the IEEE Conference on Computer Vision and Pattern Recognition (CVPR)},
  pages = {5163--5172},
  year = {2017}
}

@misc{DINOv3,
  title={{DINOv3}},
  author={Sim{\'e}oni, Oriane and Vo, Huy V. and Seitzer, Maximilian and Baldassarre, Federico and Oquab, Maxime and Jose, Cijo and Khalidov, Vasil and Szafraniec, Marc and Yi, Seungeun and Ramamonjisoa, Micha{\"e}l and Massa, Francisco and Haziza, Daniel and Wehrstedt, Luca and Wang, Jianyuan and Darcet, Timoth{\'e}e and Moutakanni, Th{\'e}o and Sentana, Leonel and Roberts, Claire and Vedaldi, Andrea and Tolan, Jamie and Brandt, John and Couprie, Camille and Mairal, Julien and J{\'e}gou, Herv{\'e} and Labatut, Patrick and Bojanowski, Piotr},
  year={2025},
  eprint={2508.10104},
  archivePrefix={arXiv},
  primaryClass={cs.CV},
  url={https://arxiv.org/abs/2508.10104},
}

@InProceedings{Skullptor,
      author    = {Artru, Noé and Hussain, Rukhshanda and Got, Emeline and Messier, Alexandre and Lindell, David and Dib, Abdallah},
      title     = {Skullptor: High Fidelity 3D Head Reconstruction in Seconds with Multi-View Normal Prediction},
      booktitle = {Proceedings of the IEEE/CVF Conference on Computer Vision and Pattern Recognition (CVPR)},
      year      = {2026}
}

@InProceedings{ContinuousLandmark, 
     author = {Chandran, Prashanth and Zoss, Gaspard and Gotardo, Paulo and Bradley, Derek}, 
     title = {Continuous Landmark Detection With 3D Queries}, 
     booktitle = {Proceedings of the IEEE/CVF Conference on Computer Vision and Pattern Recognition (CVPR)}, 
     month = {June}, 
     year = {2023}, 
     pages = {16858-16867} 
 }
}

\clearpage
\setcounter{page}{1}
\maketitlesupplementary

This supplementary material provides additional details and results. We first provide implementation details, including prior-network training, loss-weight maps, and FLAME fitting hyperparameters. We then describe the extended NeRSemble-style benchmark and report its quantitative results, followed by runtime breakdowns for single-image fitting and offline sequence reconstruction, qualitative multi-view reconstruction examples, and additional in-the-wild and NeRSemble comparisons. Finally, we present the analytic Jacobian derivation and CUDA kernel organization for the proposed Gauss-Newton solver.

\section{Implementation Details}

\noindent \textbf{Network training.} During prior-network training, The input image is cropped and resized to a resolution of \(512\times512\), and the network uses a DINOv3 ViT-B backbone. We use \(5\%\) of the images as a validation split and train on the remaining images with a batch size of \(12\) for \(25\) epochs. The learning rates are set to \(10^{-4}\) for the main model parameters and \(10^{-5}\) for the backbone. The model is trained for 2 days on 4 NVIDIA RTX 4090 GPUs.

\begin{figure}[h]
    \centering
    \includegraphics[width=0.6\linewidth]{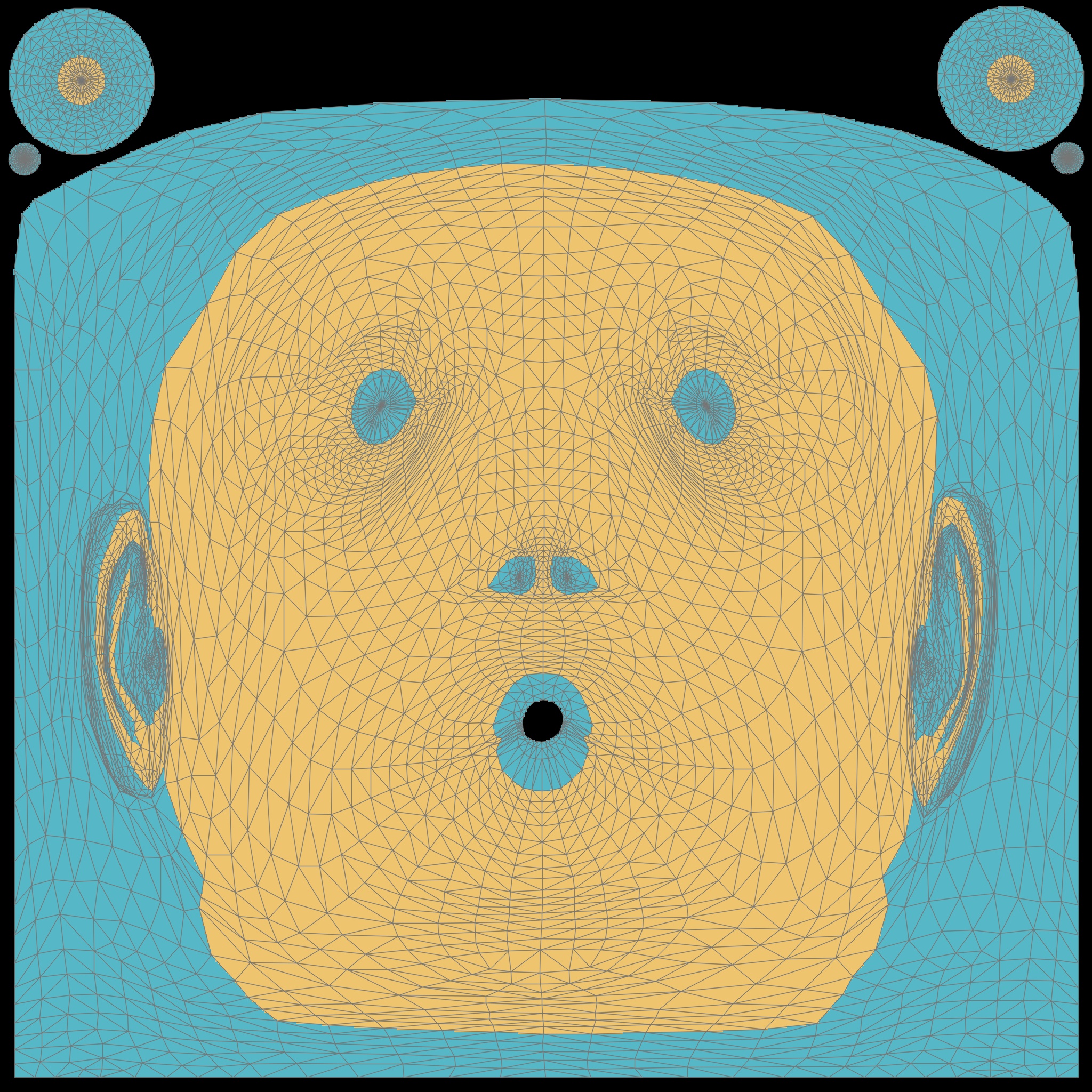}
    \caption{\textbf{Dense loss weight map.}}
    \label{fig:dense_loss_weight_map}
\end{figure}

We use a UV-space loss-weight map to define the valid supervision regions for the dense UV loss in Eq.~\eqref{eq:dense_loss} and the vertex loss in Eq.~\eqref{eq:vertex_loss}, as shown in Fig.~\ref{fig:dense_loss_weight_map}. The yellow region is assigned weight \(1.0\), the blue region is assigned weight \(0.5\), and the black region is assigned weight \(0.0\). For the dense UV loss, the weight \(w_p\) is applied at each UV pixel. For the vertex-level loss array, we sample the same weight map at the fixed FLAME vertex UV coordinates, and use the sampled value as the loss weight of each vertex \(w_i\).

\noindent \textbf{FLAME fitting.} For FLAME fitting, we set the correspondence and relative-depth weights to \(\lambda_{\mathrm{c}}=1.0,\lambda_{\mathrm{d}}=2.0\), and use \(\lambda_{\mathrm{expr}}=1\times10^{-2}\), \(\lambda_{\mathrm{id}}=3\times10^{-2}\), and \(\lambda_{\mathrm{pose}}=1\times10^{-2}\). The damping coefficient of Gauss-Newton is set to \(10^{-3}\). For single-image fitting, we first run a pose-initialization stage for \(5\) steps, where the Gauss-Newton damping is increased to \(0.5\) to avoid unstable updates of the global pose. We then alternate between optimizing the dynamic parameters \(\mathbf{x}\) and the identity parameters \(\beta\) for \(15\) steps. For offline sequence reconstruction, we optimize \(\mathbf{x}_t\) for \(10\) steps at each frame, use a keyframe buffer of \(32\) frames, and alternate between the register pass and the track pass for \(3\) rounds. For online tracking, we use a keyframe buffer of \(16\) frames, check for new keyframes every \(5\) frames, set the head-rotation novelty threshold to \(0.3 \mathrm{rad}\), and use an identity-refinement budget of \(4\) group-descent iterations.

\section{Extended NeRSemble-Style Benchmark}
We provide the full protocol and detailed results of the extended NeRSemble-style benchmark referenced in the main paper. The reproduced benchmark serves two purposes. First, it evaluates more data than the official benchmark, with \(1{,}150\) scans compared with \(391\) official scans. Second, because we control the evaluation set and intermediate outputs, it allows us to conduct qualitative comparisons and ablation studies under the same protocol.

\noindent \textbf{Protocol.} We select \(20\) NeRSemble~\cite{NeRSemble} subjects that do not appear in our training set or the \(391\) official scans. For each subject, we estimate expression coefficients using SMIRK~\cite{SMIRK} and perform farthest point sampling in expression space to select around \(60\) frames with diverse expressions. For each selected frame, we evaluate monocular reconstruction from three fixed camera views: left, center, and right. For computational efficiency, we run COLMAP on images downsampled by a factor of four, with resolution \(550\times802\), and use the resulting point clouds as reference geometry. Following Pixel3DMM~\cite{Pixel3DMM}, we additionally report recall at \(2.5\,\mathrm{mm}\) (\(R^{2.5}\)), defined as the percentage of reference points whose nearest predicted mesh surface lies within \(2.5\,\mathrm{mm}\). Since TokenFace~\cite{TokenFace} and FlowFace~\cite{FlowFace} are not publicly available, they are not included in this extended benchmark.

\begin{figure*}
    \centering
    \includegraphics[width=1.0\linewidth]{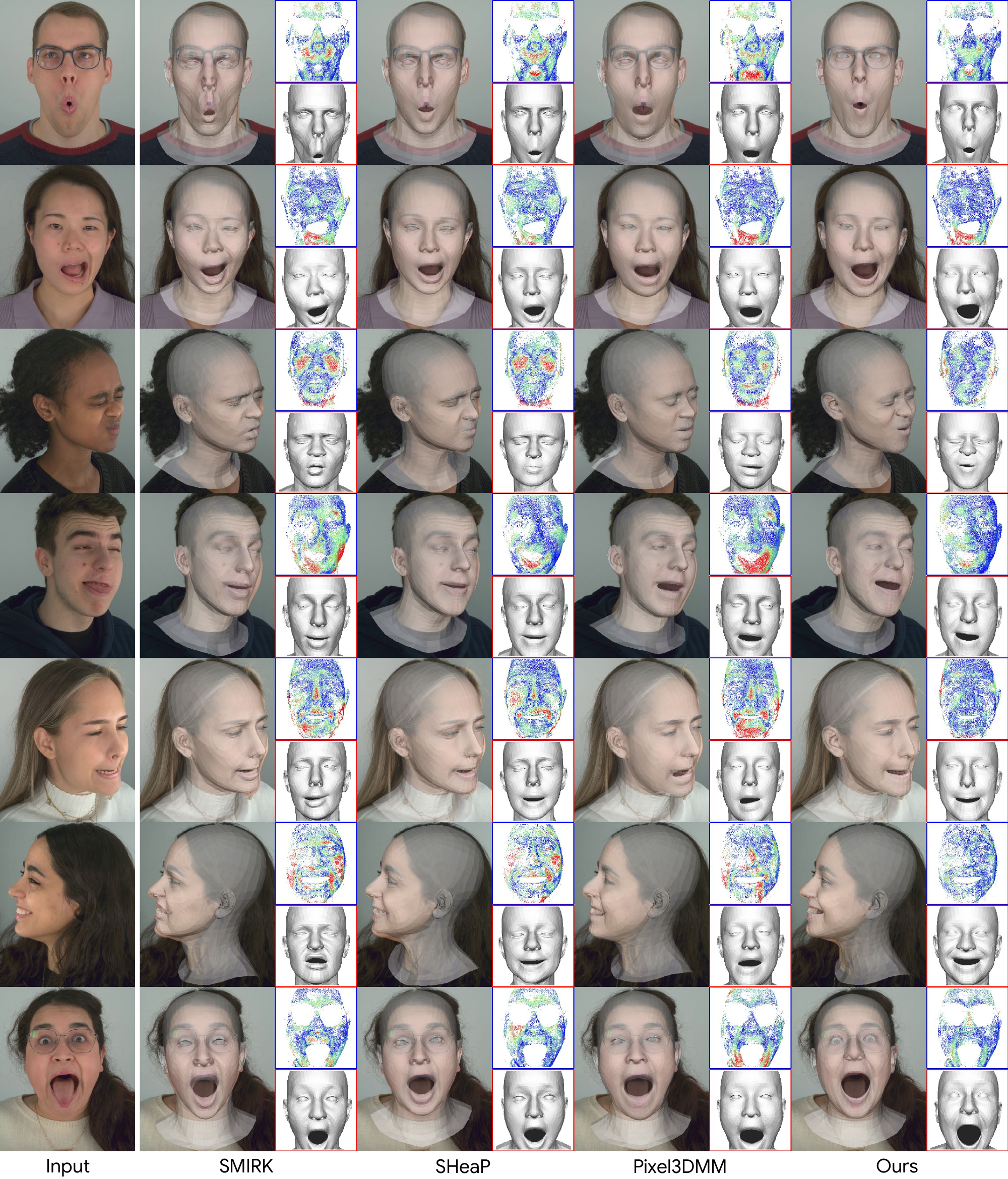}
    \caption{\textbf{Additional qualitative comparisons on the extended NeRSemble SVFR benchmark.} Our method produces consistently more faithful reconstructions than the compared baselines.}
    \label{fig:more_comp}
\end{figure*}

\noindent \textbf{Results.} Table~\ref{tab:svfr_ours} shows that our method outperforms previous baselines on the extended reproduced benchmark. The improvement is moderate but consistent on neutral reconstruction, where our method reduces L1 from \(1.93\) to \(1.86\) compared with Pixel3DMM~\cite{Pixel3DMM} and slightly improves NC and \(R^{2.5}\). The gain is much larger on posed reconstruction: our method reduces L1 from \(2.19\) to \(1.56\) and improves \(R^{2.5}\) from \(70.5\) to \(81.4\) over Pixel3DMM. This larger posed-reconstruction gain supports our main claim that dense priors and fast nonlinear least-squares fitting are especially beneficial under large head poses and strong expressions. Additional qualitative comparisons on this benchmark are shown in Fig.~\ref{fig:more_comp}.

\begin{table}[!t]
\centering
\scriptsize
\caption{\textbf{Results on NeRSemble SVFR benchmark (Extended).} Our extended benchmark follows the NeRSemble SVFR setting~\cite{NeRSemble,Pixel3DMM}, evaluates \(1{,}150\) scans, and measures each scan under three views: left, center, and right. Our method consistently outperforms previous baselines, with a particularly large advantage on posed reconstruction.}
\begin{tabular}{lcccccc}
    \toprule
     & \multicolumn{3}{c}{Neutral} & \multicolumn{3}{c}{Posed} \\
    \cmidrule(lr){2-4} \cmidrule(lr){5-7}
    Method
    & L1$\downarrow$ & NC$\uparrow$ & R$^{2.5}\uparrow$
    & L1$\downarrow$ & NC$\uparrow$ & R$^{2.5}\uparrow$ \\
    \midrule
    MICA~\cite{MICA}
    & \underline{1.90} & 0.922 & \underline{74.0} &   -  &   -  & -    \\
    SMIRK~\cite{SMIRK}
    & 2.08 & 0.918 & 69.8 & 2.49 & 0.887 & 61.6 \\
    SHeaP~\cite{SHeaP}
    & 1.94 & 0.923 & 72.6 & 2.24 & 0.901 & 67.1 \\
    Pixel3DMM~\cite{Pixel3DMM}
    & 1.93 & \underline{0.926} & 73.6 & \underline{2.19} & \underline{0.914} & \underline{70.5} \\
    Ours    & \textbf{1.86} & \textbf{0.927} & \textbf{74.6}
            & \textbf{1.56} & \textbf{0.922} & \textbf{81.4} \\
    \bottomrule
\end{tabular}
\label{tab:svfr_ours}
\end{table}

\section{Runtime Analysis}
\begin{figure}[!t]
    \centering
    \includegraphics[width=0.7\linewidth]{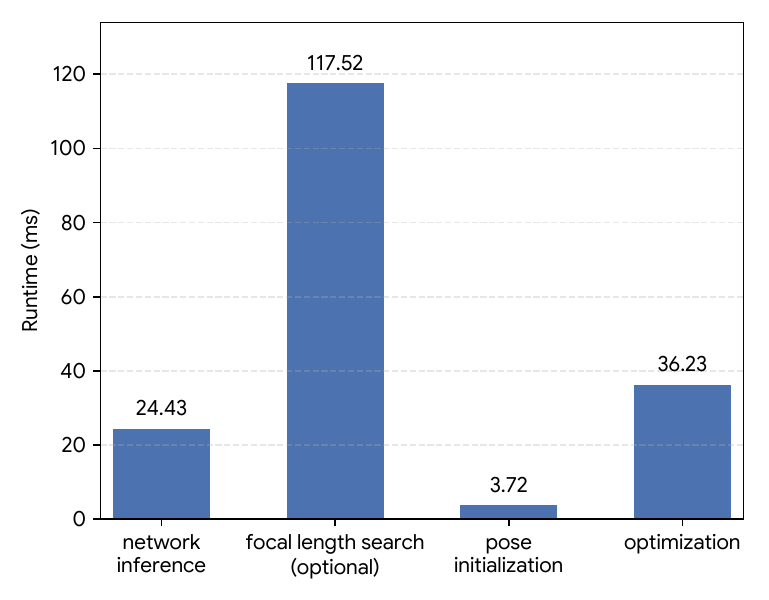}
    \caption{\textbf{Single-image fitting runtime breakdown.} The total runtime is decomposed into network inference, optional focal-length search, pose initialization, and Gauss-Newton optimization. The experiment is conducted on a single RTX 4090 GPU.}
    \label{fig:single_breakdown}
\end{figure}

For single-image fitting, the runtime breakdown on a single NVIDIA RTX 4090 GPU is shown in Fig.~\ref{fig:single_breakdown}. The full pipeline takes \(181.91\,\mathrm{ms}\) per image on average, including \(24.43\,\mathrm{ms}\) for network inference, \(117.52\,\mathrm{ms}\) for optional focal-length search, \(3.72\,\mathrm{ms}\) for pose initialization, and \(36.23\,\mathrm{ms}\) for the final Gauss-Newton optimization. The focal-length search is the most expensive component because each candidate focal length requires a short registration to evaluate the fitting energy. When camera intrinsics are known and this optional stage is skipped, the single-image runtime decreases to \(64.38\,\mathrm{ms}\), showing that the dense-prior prediction and Gauss-Newton solver are lightweight enough for practical use.

\begin{figure}[!t]
    \centering
    \includegraphics[width=1.0\linewidth]{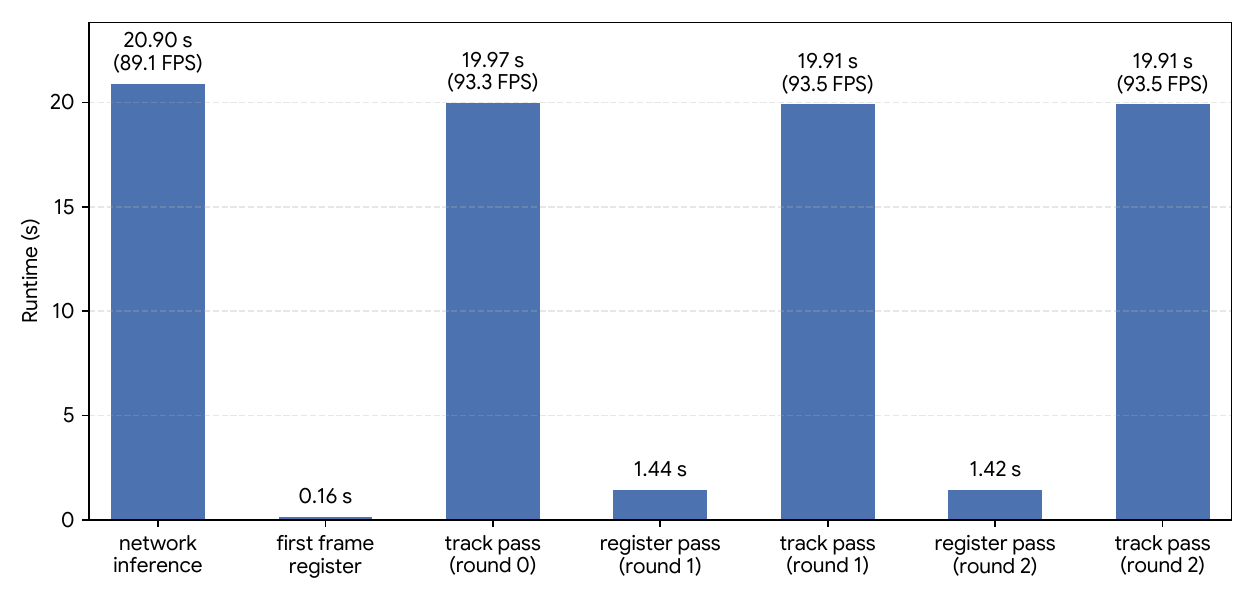}
    \caption{\textbf{Offline sequence reconstruction runtime breakdown.} The experiment is conducted on a 25.5-second video with 1862 frames at 73 FPS using a single RTX 4090 GPU, and the runtime is decomposed into network inference, first-frame registration, and the subsequent tracking and registration passes.}
    \label{fig:seq_breakdown}
\end{figure}

For offline sequence reconstruction, Fig.~\ref{fig:seq_breakdown} reports the runtime on a \(25.5\)-second video containing \(1862\) frames at \(73\) FPS, again using a single NVIDIA RTX 4090 GPU. Network inference takes \(20.90\,\mathrm{s}\) in total, corresponding to \(89.1\) FPS. The first-frame registration takes only \(0.16\,\mathrm{s}\). The dominant cost comes from the full tracking passes, which take \(19.97\,\mathrm{s}\), \(19.91\,\mathrm{s}\), and \(19.91\,\mathrm{s}\) for the three rounds, corresponding to about \(93\) FPS per pass. In contrast, the identity-refinement register passes are much cheaper, taking \(1.44\,\mathrm{s}\) and \(1.42\,\mathrm{s}\), because they are performed only on the selected keyframe buffer rather than on all video frames. These results show that our sequence pipeline is dominated by dense per-frame tracking, while keyframe-based identity refinement adds only a small overhead.

\begin{figure*}[!t]
    \centering
    \includegraphics[width=1.0\linewidth]{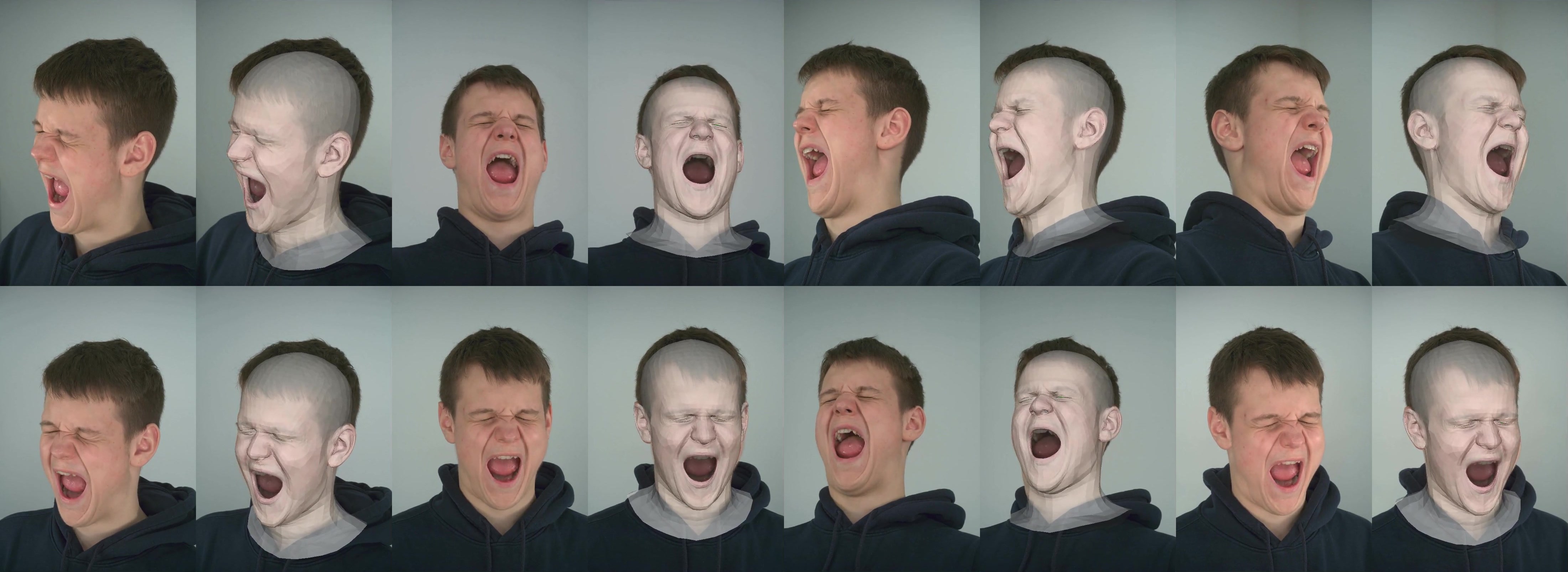}
    \includegraphics[width=1.0\linewidth]{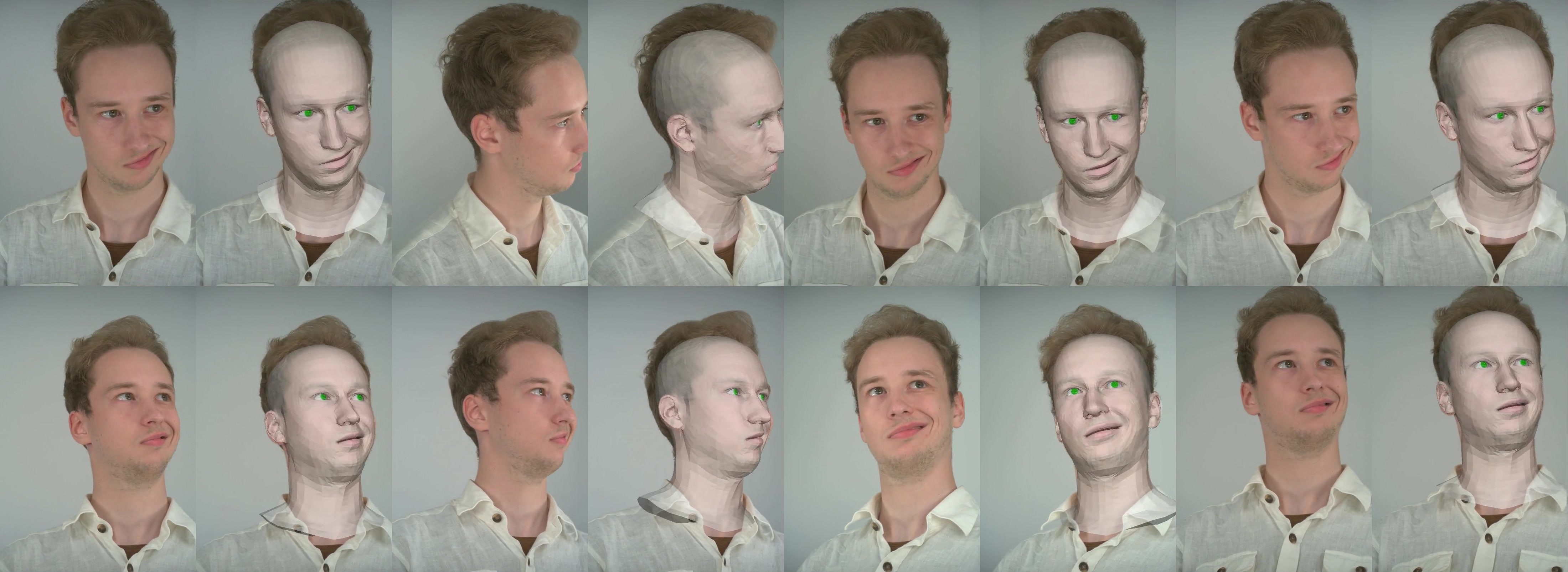}
    \caption{\textbf{Multi-view reconstruction.} We extend our fitting pipeline to calibrated multi-view sequences by jointly optimizing a single set of FLAME parameters from correspondence residuals across all views. Each pair of columns shows an input image and the fitted mesh overlay for one camera view. The results remain well aligned across large view changes and expressive facial motion.}
    \label{fig:mv_recon}
\end{figure*}

\section{Multi-view Reconstruction}
Our monocular fitting pipeline can be extended to calibrated multi-view sequences with minor modifications. Given \(N\) synchronized and calibrated cameras, we first run the prior network independently on each view to obtain per-view correspondence priors. We then jointly optimize a single set of FLAME parameters by accumulating the residuals across all views and frames. Two changes are made relative to the monocular setting. First, we disable the relative-depth prior and use only the correspondence term, because multi-view correspondence already provides sufficient geometric constraints from different viewing directions. Second, for each view we filter out the contribution of occluded vertices from the total fitting energy. We also increase the Gauss-Newton damping coefficient to \(0.1\), since the multi-view fitting energy is larger and a stronger damping helps stabilize convergence. The regularization weights for expression, pose, and identity are scaled proportionally to \(\sqrt{N}\) to balance the increased number of data residuals. Qualitative multi-view reconstruction results are shown in Fig.~\ref{fig:mv_recon}.

\begin{figure*}
    \centering
    \includegraphics[width=1.0\linewidth]{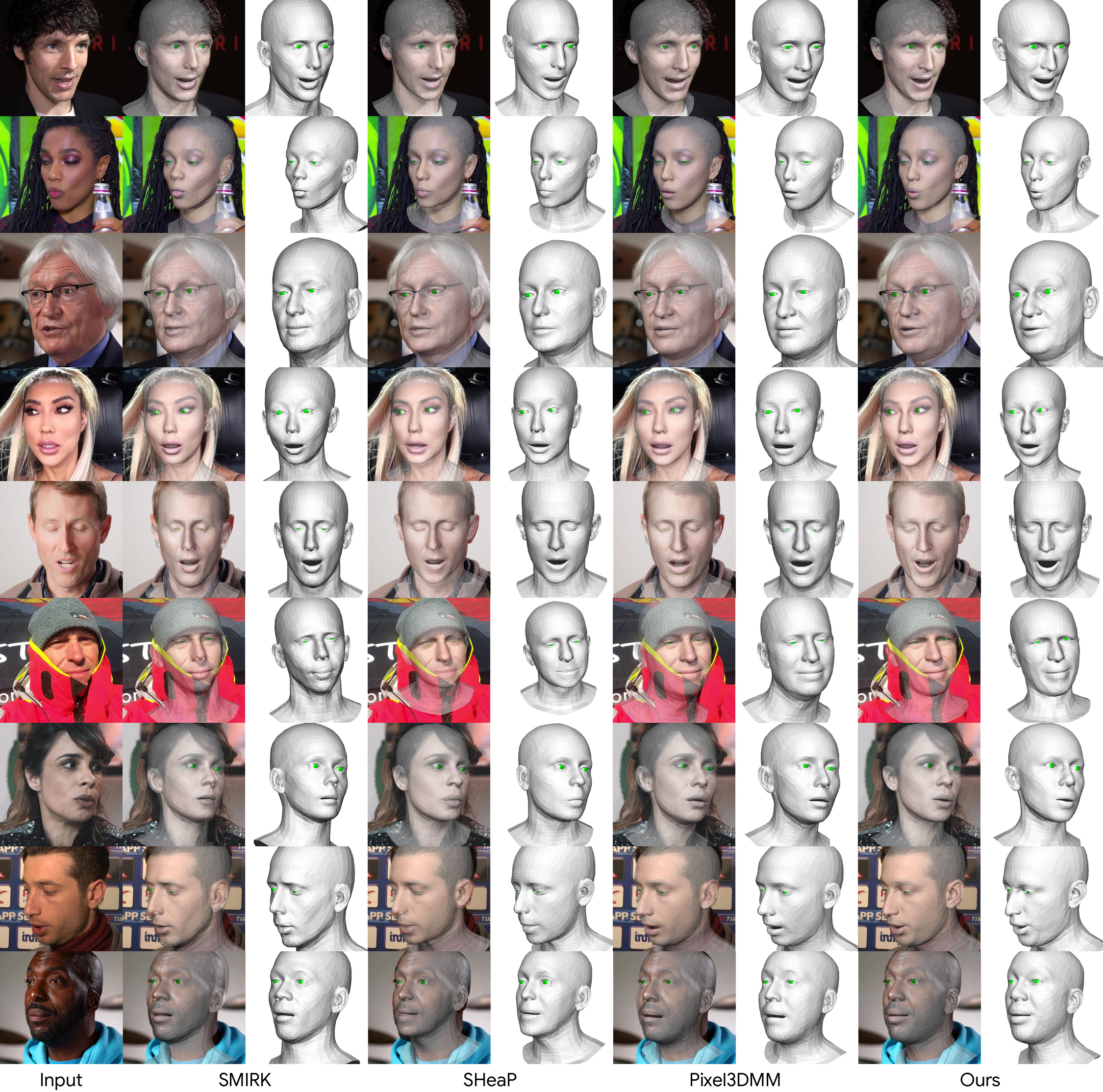}
    \caption{\textbf{Additional in-the-wild qualitative comparisons.} We compare our method with representative baselines on challenging VFHQ images~\cite{VFHQ}. The examples include strong expressions, large head poses, occlusions, and diverse appearance and lighting. Our method produces more stable FLAME alignments and more plausible facial geometry under these in-the-wild conditions.}
    \label{fig:in_the_wild_compare}
\end{figure*}

\section{Analytic Jacobian Derivation}
\subsection{FLAME Parameterization}
We use a slightly simplified FLAME parameterization for analytic Gauss-Newton fitting. As in the main paper, the identity parameters are denoted by \(\beta\in\mathbb{R}^{300}\), and the dynamic parameters of frame \(t\) are denoted as
\begin{equation}
\mathbf{x}_t
=
[
\boldsymbol{\psi}_t,\,
\boldsymbol{\theta}_t,\,
\boldsymbol{\phi}_t,\,
\mathbf{t}_t
]
\in\mathbb{R}^{118},
\end{equation}
where \(\boldsymbol{\psi}_t\in\mathbb{R}^{100}\) denotes expression, \(\boldsymbol{\theta}_t\in\mathbb{R}^{3}\) the global rotation, \(\boldsymbol{\phi}_t\in\mathbb{R}^{12}\) the local articulation parameters, and \(\mathbf{t}_t\in\mathbb{R}^{3}\) the global translation. For clarity of derivation, we omit the frame subscript in the remainder of this section.

We write the five FLAME joint rotations as \(\boldsymbol{\omega}_{k}\in\mathbb{R}^{3}\), \(k=0,\ldots,4\), where \(\boldsymbol{\omega}_{0}=\boldsymbol{\theta}\) and the remaining rotations form \(\boldsymbol{\phi}\). Let \(N=5023\) be the number of FLAME vertices and \(K=5\) the number of joints. The identity-dependent canonical vertex is
\begin{equation}
\mathbf{v}^{c}_{i}(\beta)
=
\bar{\mathbf{v}}_{i}
+
\sum_{s=1}^{300}
\beta_s \mathbf{S}_{s,i},
\end{equation}
where \(\bar{\mathbf{v}}_i\) is the template vertex and \(\mathbf{S}_{s,i}\) is the \(s\)-th identity basis displacement. Expression deformation is added by
\begin{equation}
\tilde{\mathbf{v}}_{i}
=
\mathbf{v}^{c}_{i}(\beta)
+
\sum_{l=1}^{100}
\psi_l \mathbf{E}_{l,i},
\end{equation}
where \(\mathbf{E}_{l,i}\) is the \(l\)-th expression basis displacement.

Compared with the standard FLAME model, we omit pose-dependent corrective blendshapes in the fitting solver. Empirically, we found that removing these pose corrective terms has little effect on the FLAME fitting quality in our reconstruction setting, while it simplifies the analytic Jacobians because the deformed vertices no longer contain additional rotation-dependent blendshape terms.

\subsection{Expression-Independent Joint Regression}

We use an expression-independent joint regression. In standard FLAME, joints can be regressed from vertices after both identity and expression deformation. This makes the joint locations depend on expression, so expression would influence the final posed mesh both directly through vertex deformation and indirectly through the kinematic tree. Since the expression blendshapes have little effect on the regressed FLAME joint locations in our setting, we ignore this expression-to-joint dependency in the solver.

Let \(g_{k,i}\) denote the scalar joint-regression weight from vertex \(i\) to joint \(k\). We regress canonical joints from the identity-dependent canonical mesh:
\begin{equation}
\mathbf{j}^{c}_{k}(\beta)
=
\sum_{i=1}^{N}
g_{k,i}\mathbf{v}^{c}_{i}(\beta)
=
\bar{\mathbf{j}}_{k}
+
\sum_{s=1}^{300}
\beta_s \mathbf{B}^{J}_{s,k},
\end{equation}
where
\begin{equation}
\bar{\mathbf{j}}_{k}
=
\sum_{i=1}^{N}
g_{k,i}\bar{\mathbf{v}}_{i},
\qquad
\mathbf{B}^{J}_{s,k}
=
\sum_{i=1}^{N}
g_{k,i}\mathbf{S}_{s,i}.
\end{equation}
This gives the canonical-joint Jacobians with respect to identity and expression:
\begin{equation}
\frac{\partial \mathbf{j}^{c}_{k}}{\partial \beta_s}
=
\mathbf{B}^{J}_{s,k},
\qquad
\frac{\partial \mathbf{j}^{c}_{k}}{\partial \psi_l}
=
\mathbf{0}.
\end{equation}
The first derivative is used in the identity-parameter Jacobian, while the second removes expression-to-joint terms from the dynamic-parameter Jacobian. This simplification is especially useful for the relative-depth residual, which uses the FLAME neck joint as the depth anchor.

\subsection{FLAME LBS and Jacobians}
\label{app:flame_geometry_jacobians}

\noindent \textbf{FLAME Forward LBS.}
For compactness, we write \(\mathbf v_i^c\) and \(\mathbf j_k^c\) for \(\mathbf v_i^c(\beta)\) and \(\mathbf j_k^c(\beta)\) when there is no ambiguity. Let \(\boldsymbol\omega_k\) denote the axis-angle rotation of joint \(k\), including the root rotation and the local articulation parameters. The local transform of joint \(k\) is
\begin{equation}
\mathbf T_k
=
\begin{bmatrix}
\mathbf R_k & \mathbf b_k \\
\mathbf 0^\top & 1
\end{bmatrix},
\quad
\mathbf R_k=\exp([\boldsymbol\omega_k]_\times),
\end{equation}
where \([\cdot]_\times\) denotes the skew-symmetric matrix. Let \(p(k)\) denote the parent of joint \(k\). We set \(\mathbf b_0=\mathbf j_0^c\) for the root and \(\mathbf b_k=\mathbf j_k^c-\mathbf j_{p(k)}^c\) for non-root joints. The global joint transform is obtained by forward kinematics,
\begin{equation}
\mathbf A_0=\mathbf T_0,
\quad
\mathbf A_k=\mathbf A_{p(k)}\mathbf T_k.
\end{equation}
The posed joint position is
\begin{equation}
\mathbf J_k=[\mathbf A_k]_{\mathrm{t}}+\mathbf t,
\end{equation}
where \([\cdot]_{\mathrm{t}}\) extracts the translation column of a homogeneous transform. The posed vertex is computed by linear blend skinning,
\begin{equation}
\mathbf V_i
=
\sum_{k=0}^{K-1}
w_{i,k}
\left[
\mathbf A_k
\begin{pmatrix}
\tilde{\mathbf v}_i-\mathbf j_k^c\\
1
\end{pmatrix}
\right]_{xyz}
+
\mathbf t,
\end{equation}
where \(w_{i,k}\) is the skinning weight and \([\cdot]_{xyz}\) extracts the first three coordinates of a homogeneous vector.

\noindent \textbf{Jacobian w.r.t. Dynamic Parameters.}
For dynamic-parameter updates, the identity \(\beta\) is fixed, so \(\mathbf v_i^c\) and \(\mathbf j_k^c\) are constants. The expression derivative only affects the deformed vertex \(\tilde{\mathbf v}_i\), giving
\begin{equation}
\frac{\partial \mathbf V_i}{\partial \psi_l}
=
\sum_{k=0}^{K-1}
w_{i,k}\mathbf R_k^G\mathbf E_{l,i},
\quad
\frac{\partial \mathbf J_k}{\partial \psi_l}
=
\mathbf 0,
\end{equation}
where \(\mathbf R_k^G\) is the rotation block of the global transform \(\mathbf A_k\).

For pose derivatives, we use the right Jacobian \(\mathbf J_r(\boldsymbol\omega_a)\) of \(\mathrm{SO}(3)\). Define \(\mathcal D(a)\) as the set of LBS branches whose global transform depends on joint \(a\). For vertex \(i\), the lever-arm term associated with joint \(a\) is
\begin{equation}
\mathbf p_{i,a}
=
\sum_{k\in\mathcal D(a)}
w_{i,k}
\left[
\mathbf A_a^{-1}\mathbf A_k
\begin{pmatrix}
\tilde{\mathbf v}_i-\mathbf j_k^c\\
1
\end{pmatrix}
\right]_{xyz}.
\end{equation}
The derivative of the posed vertex with respect to the axis-angle rotation of joint \(a\) is then
\begin{equation}
\frac{\partial \mathbf V_i}{\partial \boldsymbol\omega_a}
=
-\mathbf R_a^G[\mathbf p_{i,a}]_\times
\mathbf J_r(\boldsymbol\omega_a),
\end{equation}
where \([\mathbf a]_\times\mathbf b=\mathbf a\times\mathbf b\).

The joint derivative follows the same forward-kinematics structure. For each component \(\omega_{a,m}\), let \(\mathbf a_{a,m}\) be the corresponding world-space rotation axis after applying the right-Jacobian factor. If joint \(a\) is an ancestor of joint \(k\), then
\begin{equation}
\frac{\partial \mathbf J_k}{\partial \omega_{a,m}}
=
\mathbf a_{a,m}\times(\mathbf J_k-\mathbf J_a);
\end{equation}
otherwise the derivative is zero. Finally, the global translation derivative is
\begin{equation}
\frac{\partial \mathbf V_i}{\partial \mathbf t}
=
\mathbf I_3,
\quad
\frac{\partial \mathbf J_k}{\partial \mathbf t}
=
\mathbf I_3.
\end{equation}

\noindent \textbf{Jacobian w.r.t. Identity.}
For identity-parameter updates, the dynamic parameters are fixed and the active variables are the identity coefficients \(\beta_s\). The canonical vertex derivative is directly given by the identity basis,
\begin{equation}
\frac{\partial \mathbf v_i^c}{\partial \beta_s}
=
\mathbf S_{s,i}.
\end{equation}
Because the canonical joints are regressed from the identity-dependent canonical mesh, their identity derivatives are
\begin{equation}
\frac{\partial \mathbf j_k^c}{\partial \beta_s}
=
\mathbf B^J_{s,k}
=
\sum_i g_{k,i}\mathbf S_{s,i},
\end{equation}
where \(g_{k,i}\) is the joint-regression weight from vertex \(i\) to joint \(k\).

The identity coefficients affect posed joints through the local joint offsets. We denote the derivative of each global transform by
\begin{equation}
\dot{\mathbf A}_{s,k}
=
\frac{\partial \mathbf A_k}{\partial \beta_s}.
\end{equation}
For the root joint, the local offset is the canonical root joint itself, so the transform derivative contains only the identity derivative of this joint:
\begin{equation}
\dot{\mathbf A}_{s,0}
=
\begin{bmatrix}
\mathbf 0 & \mathbf B^J_{s,0}\\
\mathbf 0^\top & 0
\end{bmatrix}.
\end{equation}
For a non-root joint \(k\), identity changes the relative offset between the joint and its parent. We define
\begin{equation}
\dot{\mathbf b}_{s,k}
=
\mathbf B^J_{s,k}
-
\mathbf B^J_{s,p(k)},
\quad
\dot{\mathbf T}_{s,k}
=
\begin{bmatrix}
\mathbf 0 & \dot{\mathbf b}_{s,k}\\
\mathbf 0^\top & 0
\end{bmatrix}.
\end{equation}
The global transform derivative then follows the product rule along the kinematic tree:
\begin{equation}
\dot{\mathbf A}_{s,k}
=
\dot{\mathbf A}_{s,p(k)}\mathbf T_k
+
\mathbf A_{p(k)}\dot{\mathbf T}_{s,k}.
\end{equation}
Therefore, the posed joint Jacobian with respect to identity is
\begin{equation}
\frac{\partial \mathbf J_k}{\partial \beta_s}
=
[\dot{\mathbf A}_{s,k}]_{\mathrm{t}}.
\end{equation}
Using this joint Jacobian, the posed vertex Jacobian becomes
\begin{equation}
\frac{\partial \mathbf V_i}{\partial \beta_s}
=
\sum_{k=0}^{K-1}
w_{i,k}
\left(
\mathbf R_k^G
(\mathbf S_{s,i}-\mathbf B^J_{s,k})
+
\frac{\partial \mathbf J_k}{\partial \beta_s}
\right).
\end{equation}
In this expression, the first term accounts for the identity-dependent canonical vertex and joint offset inside each LBS branch, while the second term accounts for the identity-induced change of the posed joint transform.

\subsection{Dense-Prior Residuals and System Jacobian Assembly}

The prior network predicts a dense correspondence prior \(\hat{\mathbf u}_i\), a relative-depth prior \(\hat d_i\), and their log-variance estimates \(\hat\ell^c_i\) and \(\hat\ell^d_i\) for each sampled FLAME vertex. We first convert the predicted uncertainty into residual weights,
\begin{equation}
\alpha^c_i
=
\sqrt{\lambda_c}
\exp\left(-\frac{1}{2}\hat\ell^c_i\right),
\quad
\alpha^d_i
=
\sqrt{\lambda_d}
\exp\left(-\frac{1}{2}\hat\ell^d_i\right),
\end{equation}
where \(\lambda_c\) and \(\lambda_d\) balance the correspondence and relative-depth terms.

Given the current FLAME parameters, the model correspondence prediction is the projected vertex position,
\begin{equation}
\mathbf u_i=\Pi(\mathbf V_i),
\end{equation}
and the model relative-depth prediction is
\begin{equation}
d_i
=
\mathbf e_z^\top\mathbf R^{\mathrm{view}}
(\mathbf V_i-\mathbf J_{\mathrm{neck}}),
\end{equation}
where \(\mathbf R^{\mathrm{view}}\) is the rotation block of the camera view matrix, \(\mathbf e_z=(0,0,1)^\top\), and \(\mathbf J_{\mathrm{neck}}\) is the FLAME neck joint used as the depth anchor. The residuals are
\begin{equation}
\mathbf r^c_i
=
\alpha^c_i(\mathbf u_i-\hat{\mathbf u}_i),
\quad
r^d_i
=
\alpha^d_i(d_i-\hat d_i).
\end{equation}
For an active optimization variable \(q\), where \(q\) is either the dynamic parameters \(\mathbf x\) or the identity parameters \(\beta\), the correspondence residual Jacobian is
\begin{equation}
\frac{\partial \mathbf r^c_i}{\partial q}
=
\alpha^c_i
\frac{\partial \Pi(\mathbf V_i)}{\partial \mathbf V_i}
\frac{\partial \mathbf V_i}{\partial q}.
\end{equation}
Let
\begin{equation}
\mathbf h_i
=
\mathbf P
\begin{pmatrix}
\mathbf V_i\\
1
\end{pmatrix}
=
(h_x,h_y,h_z,h_w)^\top,
\end{equation}
where \(\mathbf P\) is the view-projection matrix. The projection maps normalized device coordinates to \([0,1]^2\),
\begin{equation}
\Pi(\mathbf V_i)
=
\frac{1}{2}
\begin{pmatrix}
h_x/h_w\\
h_y/h_w
\end{pmatrix}
+
\frac{1}{2}
\begin{pmatrix}
1\\
1
\end{pmatrix}.
\end{equation}
Since the projected coordinate is mapped to \([0,1]^2\), the projection Jacobian is
\begin{equation}
\frac{\partial \Pi(\mathbf V_i)}{\partial \mathbf V_i}
=
\frac{1}{2}
\begin{bmatrix}
\frac{h_w\mathbf p_x^\top-h_x\mathbf p_w^\top}{h_w^2}\\
\frac{h_w\mathbf p_y^\top-h_y\mathbf p_w^\top}{h_w^2}
\end{bmatrix},
\end{equation}
where \(\mathbf p_x\), \(\mathbf p_y\), and \(\mathbf p_w\) are the first three entries of the corresponding rows of \(\mathbf P\).

The relative-depth residual Jacobian is
\begin{equation}
\frac{\partial r^d_i}{\partial q}
=
\alpha^d_i
\mathbf e_z^\top\mathbf R^{\mathrm{view}}
\left(
\frac{\partial \mathbf V_i}{\partial q}
-
\frac{\partial \mathbf J_{\mathrm{neck}}}{\partial q}
\right).
\end{equation}
The required vertex and joint Jacobians are exactly those derived in Section~\ref{app:flame_geometry_jacobians}.

The full Gauss-Newton system is assembled by stacking the correspondence and relative-depth residual rows over all sampled vertices. We then append regularization rows for the active block. For dynamic-parameter updates, the regularization residual is
\begin{equation}
\mathbf r^{\mathbf x}_{\mathrm{reg}}
=
\left[
\sqrt{\lambda_{\mathrm{expr}}}\boldsymbol\psi,
\sqrt{\lambda_{\mathrm{pose}}}\boldsymbol\phi
\right],
\end{equation}
and for identity-parameter updates it is
\begin{equation}
\mathbf r^\beta_{\mathrm{reg}}
=
\sqrt{\lambda_{\mathrm{id}}}\beta.
\end{equation}

\subsection{CUDA Kernel Organization}

We implement the Gauss-Newton solver as a set of fused CUDA kernels that run entirely on the GPU, avoiding CPU--GPU synchronization within each iteration. At each iteration, a first kernel applies the identity blendshapes to the FLAME template and regresses canonical joint positions. A second kernel then traverses the kinematic tree, computes forward LBS, and simultaneously evaluates the analytic Jacobians of posed vertices and joints with respect to the active variables; separate kernels are used for dynamic-parameter and identity-parameter updates because the two variable blocks have different Jacobian structures. A third kernel projects each vertex with the camera model, computes the correspondence and relative-depth residuals, applies the predicted confidence weights, and writes the weighted Jacobian rows together with the regularization terms. Finally, the normal equation \(\mathbf{J}^\top\mathbf{J}\,\Delta=-\mathbf{J}^\top\mathbf{r}\) is formed via cuBLAS and solved by Cholesky factorization via cuSOLVER. Because the number of unknowns is small (118 for dynamic parameters, 300 for identity parameters) relative to the number of residuals (\(3\times5023\) vertices), the normal-equation approach reduces the linear solve to a small symmetric positive-definite system. All intermediate buffers are pre-allocated at initialization, so the per-iteration cost is dominated by the geometry and assembly kernels.

\end{document}